\documentclass[format=acmsmall,
review=false, screen=true, authorversion=true]{acmart}

\usepackage[british]{babel}
\usepackage[utf8]{inputenc}
\usepackage[T1]{fontenc}
\usepackage{graphicx,tabularx}
\usepackage{multirow,multicol}
\usepackage[tight]{units}
\usepackage{enumitem}
\usepackage{pdflscape,afterpage}
\usepackage[misc]{ifsym}
\usepackage{microtype}
\usepackage{booktabs}
\usepackage{xspace}
\usepackage{tikz,xcolor,rotating}
\usepackage{algpseudocode}
\usepackage[ruled]{algorithm2e}

\usepackage{fancyhdr}

\SetAlFnt{\small}
\SetAlCapFnt{\small}
\SetAlCapNameFnt{\small}
\SetAlCapHSkip{0pt}
\IncMargin{-\parindent}

\DeclareMathOperator{\tr}{tr} 
\DeclareMathOperator{\rank}{rank} 
\DeclareMathOperator{\diag}{diag} 
\DeclareMathOperator{\argmin}{argmin} 

\newcommand{\gC}{\ensuremath{C}} 
\newcommand{\gCRIT}{\ensuremath{\mathcal{J}}} 
\newcommand{\gCRITf}[1]{\ensuremath{\gCRIT\left(#1\right)}} 
\newcommand{\gD}{\ensuremath{D}} 
\newcommand{\gDELTA}{\ensuremath{\gB{\Delta}}} 
\newcommand{\gDELTAccH}[2]{\ensuremath{\gH{\delta}\left(#1,#2\right)}} 
\newcommand{\gG}{\ensuremath{\mathcal{G}}} 
\newcommand{\gGn}[1]{\ensuremath{\gB{g}_{#1}}} 
\newcommand{\gGnH}[1]{\ensuremath{\gH{\gB{g}}_{#1}}} 
\newcommand{\gGnc}[2]{\ensuremath{\gGn{#1}^{(#2)}}} 
\newcommand{\gGAMMAjt}[2]{\ensuremath{\gamma_{#1}\left(#2\right)}} 
\newcommand{\gI}{\ensuremath{\gB{I}}} 
\newcommand{\gIc}[1]{\ensuremath{\mathcal{I}_{#1}}} 
\newcommand{\gJ}{\ensuremath{J}} 
\newcommand{\gLAG}[2]{\ensuremath{\mathcal{L}\left(#1,#2\right)}} 
\newcommand{\gLAMBDA}{\ensuremath{\gB{\Lambda}}} 
\newcommand{\gLAMBDAd}[1]{\ensuremath{\lambda_{#1}}} 
\newcommand{\gLAMBDAn}[1]{\ensuremath{\ell_{#1}}} 
\newcommand{\gM}{\ensuremath{\mu}} 
\newcommand{\gMc}[1]{\ensuremath{\gM_{#1}}} 
\newcommand{\gMcH}[1]{\ensuremath{\gH{\gM}_{#1}}} 
\newcommand{\gN}{\ensuremath{N}} 
\newcommand{\gNc}[1]{\ensuremath{\gN_{#1}}} 
\newcommand{\gPHI}{\ensuremath{\gB{\Phi}}} 
\newcommand{\gPHId}[1]{\ensuremath{\gB{f}_{#1}}} 
\newcommand{\gPSI}{\ensuremath{\gB{\Psi}}} 
\newcommand{\gSIGMAc}[1]{\ensuremath{\gB{\Sigma}_{#1}}} 
\newcommand{\gSIGMAcH}[1]{\ensuremath{\gB{\gH{\Sigma}}_{#1}}} 
\newcommand{\gSIGMAb}{\ensuremath{\gSIGMAc{\mathrm{B}}}} 
\newcommand{\gSIGMAw}{\ensuremath{\gSIGMAc{\mathrm{W}}}} 
\newcommand{\gSIGMAt}{\ensuremath{\gSIGMAc{\mathrm{T}}}} 
\newcommand{\gSIGMAtH}{\ensuremath{\gSIGMAcH{\mathrm{T}}}} 
\newcommand{\gT}{\ensuremath{T}} 

\newcommand{\gTHETA}{\ensuremath{\gB{\Theta}}}
\newcommand{\gXI}{\ensuremath{\gB{\Xi}}}
\newcommand{\gUPSILON}{\ensuremath{\gB{\Upsilon}}}

\newcommand{\gCHI}{\ensuremath{\gB{X}}}

\newcommand{\gOMEGA}{\ensuremath{\gB{\Omega}}}

\newcommand{\gB}[1]{\ensuremath{\mathbf{#1}}} 
\newcommand{\gH}[1]{\ensuremath{\widehat{#1}}} 
\newcommand{\gO}[1]{\ensuremath{\overline{#1}}} 
\newcommand{\gL}[1]{\ensuremath{{#1}_L}} 
\newcommand{\gE}[1]{\ensuremath{{#1}_E}} 

\newcommand{\sub}[1]{\raisebox{-.4ex}{\scriptsize{#1}}}
\newcommand\etal{\textit{et~al.}\xspace}
\let\exper\textbf

\acmJournal{TOMM}
\acmYear{2017}
\acmMonth{10}
\copyrightyear{2017}

\setcopyright{acmcopyright}

\received{February 2017}
\received[revised]{August 2017}
\received[accepted]{August 2017}

\begin{document}
\hyphenation{homo-geneous hetero-geneous mocap}
\title[Gait~Recognition~from~Motion~Capture~Data]{Gait~Recognition~from~Motion~Capture~Data}

\author{Michal Balazia}
\orcid{0000-0001-7153-9984}
\author{Petr Sojka}
\orcid{0000-0002-5768-4007}
\affiliation{
\institution{Masaryk University} 
\department{Faculty of Informatics}
\streetaddress{Botanick\'a 68a} 
\postcode{602\,00} 
\city{Brno}
\country{Czech Republic}
}
\begin{abstract}
Gait recognition from motion capture data, as a pattern classification discipline, can be improved by the use of machine learning. This paper contributes to the state-of-the-art with a statistical approach for extracting robust gait features directly from raw data by a modification of Linear Discriminant Analysis with Maximum Margin Criterion. Experiments on the CMU MoCap database show that the suggested method outperforms thirteen relevant methods based on geometric features and a method to learn the features by a combination of Principal Component Analysis and Linear Discriminant Analysis. The methods are evaluated in terms of the distribution of biometric templates in respective feature spaces expressed in a number of class separability coefficients and classification metrics. Results also indicate a high portability of learned features, that means, we can learn what aspects of walk people generally differ in and extract those as general gait features. Recognizing people without needing group-specific features is convenient as particular people might not always provide annotated learning data. As a contribution to reproducible research, our evaluation framework and database have been made publicly available. This research makes motion capture technology directly applicable for human recognition.
\end{abstract}



\thanks{Authors thank to the reviewers for their detailed commentary and suggestions. The data used in this project was created with funding from NSF EIA-0196217 and was obtained from \url{http://mocap.cs.cmu.edu}~\cite{CMU03}. Our extracted database and evaluation framework are available online at \url{https://gait.fi.muni.cz} to support reproducibility of results.}

\maketitle

\pagestyle{plain}
\thispagestyle{fancy}
\fancyhead[L]{
ACM Transactions on Multimedia Computing, Communications, and Applications\\
special issue on Representation, Analysis and Recognition of 3D Humans, preprint}

\section{Introduction}
\label{intro}

From the surveillance perspective, gait pattern biometrics is appealing because it can be performed at a distance and without body-invasive equipment or the need for the subject's cooperation. This allows data acquisition without a subject's consent. As the data are collected with a high participation rate and the subjects are not expected to claim their identities, the trait is employed for identification rather than for authentication.

\begin{figure}[tb]
\centering
\includegraphics[width=\textwidth]{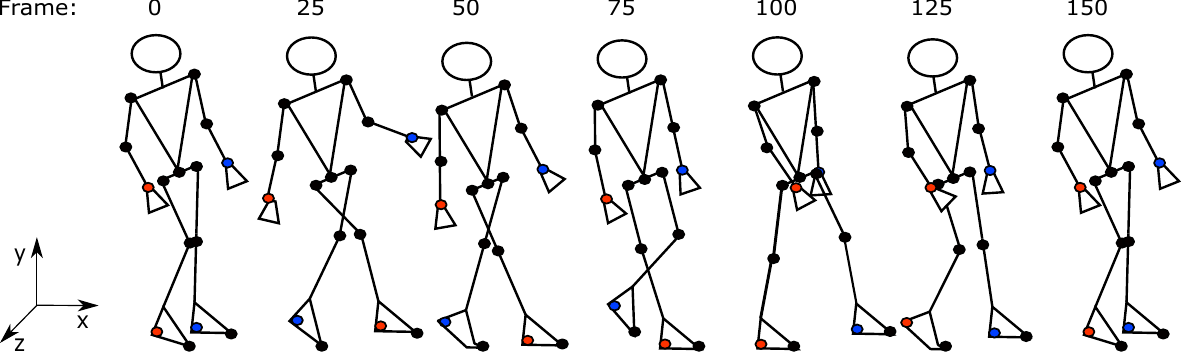}
\caption{Motion capture data. Skeleton is represented by a stick figure of 31~joints (only 15~are drawn here). Seven selected video frames of a walk sequence contain 3D spatial coordinates of each joint in time.}
\label{f1}
\end{figure}

Motion capture technology acquires video clips of individuals and generates structured motion data. The format maintains an overall structure of the human body and holds estimated 3D positions of the main anatomical landmarks as the person moves. These so-called motion capture data (MoCap) can be collected online by RGB-D sensors such as Microsoft Kinect, Asus Xtion or Vicon. To visualize motion capture data (see Figure~\ref{f1}), a simplified stick figure representing the human skeleton (a graph of joints connected by bones) can be automatically recovered from the values of body point spatial coordinates. With recent rapid improvements in MoCap sensor accuracy, we foresee an affordable MoCap technology~\cite{HALP13} that can identify people from MoCap data.

The goal of this work is to present a method for learning robust gait features from raw MoCap data. A collection of extracted features builds a gait template that serves as the walker's signature. Templates are stored in a central database. Recognition of a person involves capturing their walk sample, extracting gait features to compose a template, and finally querying this database for a~set of similar templates to report the most likely identity. The similarity of the two templates is expressed in a single number computed by a similarity\slash distance function.

Related work is outlined in Section~\ref{rel} and our methods are described in detail in Section~\ref{meth}. We provide a thorough evaluation of a number of competitive MoCap-based gait recognition methods on a benchmark database and framework described in Section~\ref{eval-data}. Two setups of data separation into learning and evaluation parts are presented in~\ref{eval-setup}, together with several metrics used for thorough evaluation. Results are presented and discussed in Section~\ref{eval-res}.

\section{Related Work}
\label{rel}

Human gait has been an active subject of study in physical medicine (detection of gait abnormalities~\cite{CPF15,DWDBDP16} and disorders caused by stroke or cerebral palsy~\cite{KB06}), sport (gait regulation~\cite{WB11}), or sociology (age and gender classification~\cite{MMY10} or attractiveness evaluation~\cite{HKP14}) for a long time. Inspired by medical studies of Murray~\cite{M67}, a psychological research of Johansson~\cite{J73} and by Cutting and Kozlowski~\cite{CK77} conducted experiments about how participants are able to recognize pedestrians from simply observing the 2D gait pattern generated by light bulbs attached to several joints over their body. These experiments proved that the gait is personally unique and can be potentially used for biometric recognition. The main challenges from the perspective of biometric security~\cite{BPM09} are (1) to detect gait cycles in long video footage; (2) to recognize registered participants by their biometric samples; and (3) to retrieve relevant biometric samples from large databases.

Many research groups investigate the discrimination power of gait pattern and develop models of human walk for applications in the automatic recognition from MoCap data of walking people. A number of MoCap-based gait recognition methods have been introduced in the past few years and new ones continue to emerge. In order to move forward with the wealth and scope of competitive research, it is necessary to compare the innovative approaches with the state-of-the-art and evaluate them against established evaluation metrics on a benchmark database. New frameworks and databases have been developed recently~\cite{CK13,KTTEF15}.

Over the past few years, most of the introduced gait features are on a geometric basis. They typically combine static body parameters (bone lengths, person's height) with dynamic gait features such as step length, walk speed, joint angles and inter-joint distances, along with various statistics (mean, standard deviation or local\slash global extremes) of their signals. We are particularly focused on the dynamic parameters. By our definition, gait is a dynamic behavioral trait. Static body parameters are not associated with gait, however, they can be used as another biometric.

What follows is a detailed specification of the thirteen gait features extraction methods that we have reviewed in our work to date. Since the idea behind each method has some potential, we have implemented each of them for direct comparison.

\def\method #1 {\item[\textbullet] \textbf{#1}\space}
\begin{itemize}
\method AhmedF by Ahmed~\etal~\cite{APG15} chooses 20~joint relative distance signals and 16~joint relative angle signals across the whole body, compared using the Dynamic Time Warping (DTW).
\method AhmedM by Ahmed~\etal~\cite{AAS14} extracts the mean, standard deviation and skew during one gait cycle of horizontal distances (projected on the Z~axis) between feet, knees, wrists and shoulders, and mean and standard deviation during one gait cycle of vertical distances (Y~coordinates) of head, wrists, shoulders, knees and feet, and finally the mean area during one gait cycle of the triangle of root and two feet.
\method AliS by Ali~\etal~\cite{AWLSWZ16} measures the mean areas during one gait cycle of lower limb triangles.
\method AnderssonVO by Andersson and Araujo~\cite{AA15} calculates gait attributes as mean and standard deviation during one gait cycle of local extremes of the signals of lower body angles, step length as a maximum of feet distance, stride length as a length of two steps, cycle time and velocity as a ratio of stride length and cycle time. In addition, they extract the mean and standard deviation during one gait cycle of each bone length, and height as the sum of the bone lengths between head and root plus the averages of the bone lengths between root and both feet.
\method BallA by Ball~\etal~\cite{BRRV12} measures mean, standard deviation and maximum during one gait cycle of lower limb angle pairs: upper leg relative to the Y~axis, lower leg relative to the upper leg, and the foot relative to the Z~axis.
\method DikovskiB by Dikovski~\etal~\cite{DMG14} selects the mean during one gait cycle of step length, height, all bone lengths, then mean, standard deviation, minimum, maximum and mean difference of subsequent frames during one gait cycle of all major joint angles, and the angle between the lines of the shoulder joints and the hip joints.
\method JiangS by Jiang~\etal~\cite{JWZS15} measures angle signals between the Y~axis and four major lower body (thigh and calf) bones. The signals are compared using the DTW.
\method KrzeszowskiT by Krzeszowski~\etal~\cite{KSKJW14} observes the signals of rotations of eight major bones (humerus, ulna, thigh and calf) around all three axes, the person's height and step length. These signals are compared using the DTW distance function.
\method KwolekB by Kwolek~\etal~\cite{KKMJ14} processes signals of bone angles around all axes, the person's height and step length. The gait cycles are normalized to 30~frames.
\method NareshKumarMS by Naresh Kumar and Venkatesh Babu~\cite{NV12} is an interesting approach that extracts all joint trajectories around all three axes and compares gait templates by a distance function of their covariance matrices.
\method PreisJ by Preis~\etal~\cite{PKWL12} takes height, length of legs, torso, both lower legs, both thighs, both upper arms, both forearms, step length and speed.
\method SedmidubskyJ by Sedmidubsky~\etal~\cite{SVBZ12} concludes that only the two shoulder-hand signals are discriminatory enough to be used for recognition. These temporal data are compared using the DTW distance function.
\method SinhaA by Sinha~\etal~\cite{SCB13} combines all features of BallA and PreisJ with mean areas during one gait cycle of upper body and lower body, then mean, standard deviation and maximum distances during one gait cycle between the centroid of the upper body polygon and the centroids of four limb polygons.
\end{itemize}

The aforementioned features are schematic and human-interpretable, which is convenient for visualizations and for intuitive understanding of the models, but unnecessary for automatic gait recognition. Instead, to refrain from ad-hoc schemes and to explore beyond the limits of human interpretability, we prefer learning gait features that maximally separate the identity classes in the feature space. The features calculated by statistical observation of large amounts of data are expected to have a much higher discriminative potential, which will be the subject of experimental evaluation in Section~\ref{eval}.

Methods for 2D~gait recognition extensively use machine learning models for extracting gait features, such as principal component analysis and multi-scale shape analysis~\cite{CT15}, genetic algorithms and kernel principal component analysis~\cite{TBLH15}, radial basis function neural networks~\cite{ZW14}, or convolutional neural networks~\cite{CMGP16}. All of these and many other models can be reasonably utilized in 3D~gait recognition as well. The following section provides a scheme for learning the features directly from raw data by (i)~a~modification of Fisher's Linear Discriminant Analysis with Maximum Margin Criterion and (ii)~a~combination of Principal Component Analysis and Linear Discriminant Analysis.

\section{Learning Gait Features}
\label{meth}

In statistical pattern recognition, reducing space dimensionality is a common technique to overcome class estimation problems. Classes are discriminated by projecting high-dimensional input data onto low-dimensional sub-spaces by linear transformations with the goal of maximizing class separability. We are interested in finding an optimal feature space where a gait template is close to those of the same walker and far from those of different walkers.

Let the model of a human body have $\gJ$ joints and all $\gL{\gN}$ learning samples of $\gL{\gC}$ walkers be linearly normalized to their average length~$\gT$. Labeled learning data in a~sample (measurement) space have the form $\gL{\gG}=\left\{\left(\gGn{n},\gLAMBDAn{n}\right)\right\}_{n=1}^{\gL{\gN}}$ where
\begin{equation}
\gGn{n}=\left[[\gGAMMAjt{1}{1}\,\cdots\,\gGAMMAjt{\gJ}{1}]\,\cdots\,[\gGAMMAjt{1}{\gT}\,\cdots\,\gGAMMAjt{\gJ}{\gT}]\right]^\top
\end{equation}
is a gait sample (one gait cycle) in which $\gGAMMAjt{j}{t}\in\mathbb{R}^3$ are 3D spatial coordinates of joint $j\in\left\{1,\ldots,\gJ\right\}$ at time $t\in\left\{1,\ldots,\gT\right\}$ normalized with respect to the person's position and walk direction. See that $\gL{\gG}$ has dimensionality $\gD=3\gJ\gT$. Each learning sample falls strictly into one of the learning identity classes $\left\{\gIc{c}\right\}_{c=1}^{\gL{\gC}}$ labeled by $\gLAMBDAn{n}$. A~class $\gIc{c}\subseteq\gL{\gG}$ has $\gNc{c}$ samples. The classes are complete and mutually exclusive. We say that samples $\left(\gGn{n},\gLAMBDAn{n}\right)$ and $\left(\gGn{n'},\gLAMBDAn{n'}\right)$ share a~common walker if and only if they belong to the same class, i.e., $\left(\gGn{n},\gLAMBDAn{n}\right),\left(\gGn{n'},\gLAMBDAn{n'}\right)\in\gIc{c}\Leftrightarrow\gLAMBDAn{n}=\gLAMBDAn{n'}$. For the whole labeled data, we denote the between- and within-class and total scatter matrices
\begin{equation}
\begin{split}
\gSIGMAb & =\sum_{c=1}^{\gL{\gC}}\left(\gMc{c}-\gM\right)\left(\gMc{c}-\gM\right)^\top\\
\gSIGMAw & =\sum_{c=1}^{\gL{\gC}}\frac{1}{\gNc{c}}\sum_{n=1}^{\gNc{c}}\left(\gGnc{n}{c}-\gMc{c}\right)\left(\gGnc{n}{c}-\gMc{c}\right)^\top\\
\gSIGMAt & =\sum_{c=1}^{\gL{\gC}}\frac{1}{\gNc{c}}\sum_{n=1}^{\gNc{c}}\left(\gGnc{n}{c}-\gM\right)\left(\gGnc{n}{c}-\gM\right)^\top=\gSIGMAb+\gSIGMAw
\end{split}
\end{equation}
where $\gGnc{n}{c}$ denotes the $n$-th sample in class $\gIc{c}$ and $\gMc{c}$ and $\gM$ are sample means for class $\gIc{c}$ and the whole data set, respectively, that is, $\gMc{c}=\frac{1}{\gNc{c}}\sum_{n=1}^{\gNc{c}}\gGnc{n}{c}$ and $\gM=\frac{1}{\gL{\gN}}\sum_{n=1}^{\gL{\gN}}\gGn{n}$. Margin is defined as the Euclidean distance of class means minus both individual variances (traces of scatter matrices)
\begin{equation}
\gSIGMAc{c}=\frac{1}{\gNc{c}}\sum_{n=1}^{\gNc{c}}\left(\gGnc{n}{c}-\gMc{c}\right)\left(\gGnc{n}{c}-\gMc{c}\right)^\top.
\end{equation}

We measure class separability of a given feature space by a representation of the Maximum Margin Criterion (MMC)~\cite{KKS04,LJZ06} used by the Vapnik's Support Vector Machines (SVM)~\cite{V95} which maximizes the sum of $\frac{1}{2}\gL{\gC}(\gL{\gC}-1)$ between-class margins
\begin{equation}
\begin{split}
\gCRIT & =\frac{1}{2}\sum_{c,c'=1}^{\gL{\gC}}
\left(\left(\gMc{c}-\gMc{c'}\right)^\top\left(\gMc{c}-\gMc{c'}\right)-\tr\left(\gSIGMAc{c}+\gSIGMAc{c'}\right)\right)\\
& =\frac{1}{2}\sum_{c,c'=1}^{\gL{\gC}}\left(\gMc{c}-\gMc{c'}\right)^\top\left(\gMc{c}-\gMc{c'}\right)-\frac{1}{2}\sum_{c,c'=1}^{\gL{\gC}}\tr\left(\gSIGMAc{c}+\gSIGMAc{c'}\right)\\
& =\frac{1}{2}\sum_{c,c'=1}^{\gL{\gC}}\left(\gMc{c}-\gM+\gM-\gMc{c'}\right)^\top\left(\gMc{c}-\gM+\gM-\gMc{c'}\right)-\sum_{c=1}^{\gL{\gC}}\tr\left(\gSIGMAc{c}\right)\\
& =\tr\left(\sum_{c=1}^{\gL{\gC}}\left(\gMc{c}-\gM\right)\left(\gMc{c}-\gM\right)^\top\right)-\tr\left(\sum_{c=1}^{\gL{\gC}}\gSIGMAc{c}\right)\\
& =\tr\left(\gSIGMAb\right)-\tr\left(\gSIGMAw\right)=\tr\left(\gSIGMAb-\gSIGMAw\right).
\end{split}
\end{equation}
Since $\tr\left(\gSIGMAb\right)$ measures the overall variance of the class mean vectors, a large one implies that the class mean vectors scatter in a large space. On the other hand, a small $\tr\left(\gSIGMAw\right)$ implies that classes have a small spread. Thus, a large $\gCRIT$ indicates that samples are close to each other if they share a~common walker but are far from each other if they are performed by different walkers. Extracting features, that is, transforming the input data in the sample space into a feature space of higher $\gCRIT$, can be used to link new observations of walkers more successfully.

Feature extraction is given by a linear transformation (feature) matrix $\gPHI\in\mathbb{R}^{\gD\times\gH{\gD}}$ from a $\gD$-dimensional sample space $\gG=\left\{\gGn{n}\right\}_{n=1}^{\gN}$ of not necessarily labeled gait samples to a $\gH{\gD}$-dimensional feature space $\gH{\gG}=\left\{\gGnH{n}\right\}_{n=1}^{\gN}$ of gait templates where $\gH{\gD}<\gD$ and each gait sample $\gGn{n}$ is transformed into a gait template $\gGnH{n}=\gPHI^\top\gGn{n}$. The objective is to learn a transform $\gPHI$ that maximizes MMC in the feature space
\begin{equation}
\gCRITf{\gPHI}=\tr\left(\gPHI^\top\left(\gSIGMAb-\gSIGMAw\right)\gPHI\right).
\label{e2}
\end{equation}
Once the transformation is found, all measured samples are transformed into templates (in the feature space) along with the class means and covariances. The templates are compared by the Mahalanobis distance function
\begin{equation}
\gDELTAccH{\gGnH{n}}{\gGnH{n'}}=\sqrt{\left(\gGnH{n}-\gGnH{n'}\right)^\top\gSIGMAtH^{-1}\left(\gGnH{n}-\gGnH{n'}\right)}.
\label{e3}
\end{equation}

We show that a solution to the optimization problem in Equation~\eqref{e2} can be obtained by eigendecomposition of the matrix $\gSIGMAb-\gSIGMAw$. An important property to notice about the objective $\gCRITf{\gPHI}$ is that it is invariant w.r.t.\@ rescalings $\gPHI\rightarrow\alpha\gPHI$. Hence, we can always choose $\gPHI=\gPHId{1}\|\cdots\|\gPHId{\gH{\gD}}$ such that $\gPHId{\gH{d}}^\top\gPHId{\gH{d}}=1$, since it is a scalar itself. For this reason we can reduce the problem of maximizing $\gCRITf{\gPHI}$ into the constrained optimization problem
\begin{equation}
\begin{split}
\max & \enskip\sum_{\gH{d}=1}^{\gH{\gD}}\gPHId{\gH{d}}^\top\left(\gSIGMAb-\gSIGMAw\right)\gPHId{\gH{d}}\\
\mathrm{subject\,to} & \enskip\gPHId{\gH{d}}^\top\gPHId{\gH{d}}-1=0\enskip\forall\gH{d}\!\in\!\{1,\ldots,\gH{\gD}\}.
\end{split}
\end{equation}
To solve the above optimization problem, let us consider the Lagrangian
\begin{equation}
\gLAG{\gPHId{\gH{d}}}{\gLAMBDAd{\gH{d}}}=\sum_{\gH{d}=1}^{\gH{\gD}}\gPHId{\gH{d}}^\top\left(\gSIGMAb-\gSIGMAw\right)\gPHId{\gH{d}}-\gLAMBDAd{\gH{d}}\left(\gPHId{\gH{d}}^\top\gPHId{\gH{d}}-1\right)
\end{equation}
with multipliers $\gLAMBDAd{\gH{d}}$. To find the maximum, we derive it with respect to $\gPHId{\gH{d}}$ and equate it to zero
\begin{equation}
\frac{\partial\gLAG{\gPHId{\gH{d}}}{\gLAMBDAd{\gH{d}}}}{\partial\gPHId{\gH{d}}}=\left(\left(\gSIGMAb-\gSIGMAw\right)-\gLAMBDAd{\gH{d}}\gI\right)\gPHId{\gH{d}}=0
\end{equation}
which leads to
\begin{equation}
\left(\gSIGMAb-\gSIGMAw\right)\gPHId{\gH{d}}=\gLAMBDAd{\gH{d}}\gPHId{\gH{d}}
\end{equation}
where $\gLAMBDAd{\gH{d}}$ are the eigenvalues of $\gSIGMAb-\gSIGMAw$ and $\gPHId{\gH{d}}$ are the corresponding eigenvectors. Putting it all together,
\begin{equation}
\left(\gSIGMAb-\gSIGMAw\right)\gPHI=\gLAMBDA\gPHI
\end{equation}
where $\gLAMBDA=\diag\left(\gLAMBDAd{1},\ldots,\gLAMBDAd{\gH{\gD}}\right)$ is the eigenvalue matrix. Therefore,
\begin{equation}
\gCRITf{\gPHI}=\tr\left(\gPHI^\top\left(\gSIGMAb-\gSIGMAw\right)\gPHI\right)=\tr\left(\gPHI^\top\gLAMBDA\gPHI\right)=\sum_{\gH{d}=1}^{\gH{\gD}}\gPHId{\gH{d}}^\top\gLAMBDAd{\gH{d}}\gPHId{\gH{d}}=\sum_{\gH{d}=1}^{\gH{\gD}}\gLAMBDAd{\gH{d}}\gPHId{\gH{d}}^\top\gPHId{\gH{d}}=\sum_{\gH{d}=1}^{\gH{\gD}}\gLAMBDAd{\gH{d}}=\tr\left(\gLAMBDA\right)
\end{equation}
is maximized when $\gLAMBDA$ has $\gH{\gD}$ largest eigenvalues and $\gPHI$ contains the corresponding eigenvectors.

In the following we discuss how to calculate the eigenvectors of $\gSIGMAb-\gSIGMAw$ and to determine an optimal dimensionality $\gH\gD$ of the feature space. Rewrite $\gSIGMAb-\gSIGMAw=2\gSIGMAb-\gSIGMAt$. Note that the null space of $\gSIGMAt$ is a subspace of that of $\gSIGMAb$ since the null space of $\gSIGMAt$ is the common null space of $\gSIGMAb$ and $\gSIGMAw$. Thus, we can simultaneously diagonalize $\gSIGMAb$ and $\gSIGMAt$ to some $\gDELTA$ and $\gI$
\begin{equation}
\begin{split}
\gPSI^\top\gSIGMAb\gPSI & =\gDELTA\\
\gPSI^\top\gSIGMAt\gPSI & =\gI
\end{split}
\end{equation}
with the $\gD\times\rank\left(\gSIGMAt\right)$ eigenvector matrix
\begin{equation}
\gPSI=\gOMEGA\gTHETA^{-\frac{1}{2}}\gXI
\end{equation}
where $\gOMEGA$ and $\gTHETA$ are the eigenvector and corresponding eigenvalue matrices of $\gSIGMAt$, respectively, and $\gXI$ is the eigenvector matrix of $\gTHETA^{-1/2}\gOMEGA^\top\gSIGMAb\gOMEGA\gTHETA^{-1/2}$. To calculate $\gPSI$, we use a fast two-step algorithm in virtue of Singular Value Decomposition (SVD). SVD expresses a real $r \times s$ matrix $\gB{A}$ as a product $\gB{A}=\gB{U}\gB{D}\gB{V}^\top$ where $\gB{D}$ is a diagonal matrix with decreasing non-negative entries, and $\gB{U}$ and $\gB{V}$ are $r\times\min\left\{r,s\right\}$ and $s\times\min\left\{r,s\right\}$ eigenvector matrices of $\gB{A}\gB{A}^\top$ and $\gB{A}^\top\gB{A}$, respectively, and the non-vanishing entries of $\gB{D}$ are square roots of the non-zero corresponding eigenvalues of both $\gB{A}\gB{A}^\top$ and $\gB{A}^\top\gB{A}$. See that $\gSIGMAt$ and $\gSIGMAb$ can have the forms
\begin{equation}
\begin{split}
\gSIGMAt=&\enskip\gCHI\gCHI^\top\enskip\mathrm{where}\enskip\gCHI=\frac{1}{\sqrt{\gL{\gN}}}\left[\left(\gGn{1}-\gM\right)\cdots\left(\gGn{\gL{\gN}}-\gM\right)\right]\enskip\text{and}\\
\gSIGMAb=&\enskip\gUPSILON\gUPSILON^\top\enskip\text{where}\enskip\gUPSILON=\left[\left(\gMc{1}-\gM\right)\cdots\left(\gMc{\gL{\gC}}-\gM\right)\right],
\end{split}
\end{equation}
respectively. Hence, we can obtain the eigenvectors $\gOMEGA$ and the corresponding eigenvalues $\gTHETA$ of $\gSIGMAt$ through the SVD of $\gCHI$ and analogically $\gXI$ of $\gTHETA^{\nicefrac{-1}{2}}\gOMEGA^\top\gSIGMAb\gOMEGA\gTHETA^{\nicefrac{-1}{2}}$ through the SVD of $\gTHETA^{\nicefrac{-1}{2}}\gOMEGA^\top\gUPSILON$. The columns of $\gPSI$ are clearly the eigenvectors of $2\gSIGMAb-\gSIGMAt$ with the corresponding eigenvalues $2\gDELTA-\gI$. Therefore, to constitute the transform $\gPHI$ by maximizing the MMC, we should choose the eigenvectors in $\gPSI$ that correspond to the eigenvalues of at least $\nicefrac{1}{2}$ in $\gDELTA$. Note that $\gDELTA$ contains at most $\rank\left(\gSIGMAb\right)=\gL{\gC}-1$ positive eigenvalues, which gives an upper bound on the feature space dimensionality $\gH\gD$.

We found inspiration in the Fisher Linear Discriminant Analysis (LDA)~\cite{F36} that uses Fisher's criterion
\begin{equation}
\gCRITf{\gPHI_\mathrm{LDA}}=\tr\left(\frac{\gPHI_\mathrm{LDA}^\top\gSIGMAb\gPHI_\mathrm{LDA}}{\gPHI_\mathrm{LDA}^\top\gSIGMAw\gPHI_\mathrm{LDA}}\right).
\end{equation}
However, since the rank of $\gSIGMAw$ is at most $\gL{\gN}-\gL{\gC}$, it is a singular (non-invertible) matrix if $\gL{\gN}$ is less than $\gD+\gL{\gC}$, or, analogously might be unstable if $\gL{\gN}\ll\gD$. Small sample size is a substantial difficulty as it is necessary to calculate $\gSIGMAw^{-1}$. To alleviate this, the measured data can be first projected to a lower dimensional space using Principal Component Analysis (PCA), resulting in a two-stage PCA+LDA feature extraction technique~\cite{BHK97} originally introduced for face recognition
\begin{equation}
\begin{split}
\gCRITf{\gPHI_\mathrm{PCA}}=&\tr\left(\gPHI_\mathrm{PCA}^\top\gSIGMAt\gPHI_\mathrm{PCA}\right)\\
\gCRITf{\gPHI_\mathrm{LDA}}=&\tr\left(\frac{\gPHI_\mathrm{LDA}^\top\gPHI_\mathrm{PCA}^\top\gSIGMAb\gPHI_\mathrm{PCA}\gPHI_\mathrm{LDA}}{\gPHI_\mathrm{LDA}^\top\gPHI_\mathrm{PCA}^\top\gSIGMAw\gPHI_\mathrm{PCA}\gPHI_\mathrm{LDA}}\right)
\end{split}
\end{equation}
and the final transform is $\gPHI=\gPHI_\mathrm{PCA}\gPHI_\mathrm{LDA}$. Given that there are $\gO{\gD}$ principal components, then regardless of the dimensionality $\gD$ there are at least $\gO{\gD}+1$ independent data points. Thus, if the $\gO{\gD}\times\gO{\gD}$ matrix $\gPHI_\mathrm{PCA}^\top\gSIGMAw\gPHI_\mathrm{PCA}$ is estimated from $\gL{\gN}-\gL{\gC}$ independent observations and providing the $\gL{\gC}\leq\gO{\gD}\leq\gL{\gN}-\gL{\gC}$, we can always invert $\gPHI_\mathrm{PCA}^\top\gSIGMAw\gPHI_\mathrm{PCA}$ and in this way obtain the LDA estimate. Note that this method is sub-optimal for multi-class problems~\cite{LDH01} as PCA keeps at most $\gL{\gN}-\gL{\gC}$ principal components whereas at least $\gL{\gN}-1$ of them are necessary in order not to lose information. PCA+LDA in this form has been used for silhouette-based (2D) gait recognition by Su~\etal~\cite{SLC09} and is included in our experiments with MoCap~(3D).

On given labeled learning data $\gL{\gG}$, Algorithm~\ref{a1} and Algorithm~\ref{a2}~\cite{BS16a,BS16b} provided below are efficient ways of learning the transforms $\gPHI$ for MMC and PCA+LDA, respectively.

\begin{algorithm}[tb]
\caption{LearnTransformationMatrixMMC$\left(\gL{\gG}\right)$}
\label{a1}
\begin{algorithmic}[1]
  \State split $\gL{\gG}=\left\{\left(\gGn{n},\gLAMBDAn{n}\right)\right\}_{n=1}^{\gL{\gN}}$ into classes $\left\{\gIc{c}\right\}_{c=1}^{\gL{\gC}}$ of $\gNc{c}=\left|\gIc{c}\right|$ samples
  \State compute overall mean $\gM=\frac{1}{\gL{\gN}}\sum_{n=1}^{\gL{\gN}}\gGn{n}$ and individual class means $\gMc{c}=\frac{1}{\gNc{c}}\sum_{n=1}^{\gNc{c}}\gGnc{n}{c}$
  \State compute $\gSIGMAb=\sum_{c=1}^{\gL{\gC}}\left(\gMc{c}-\gM\right)\left(\gMc{c}-\gM\right)^\top$
  \State compute $\gCHI=\frac{1}{\sqrt{\gL{\gN}}}\left[\left(\gGn{1}-\gM\right)\cdots\left(\gGn{\gL{\gN}}-\gM\right)\right]$
  \State compute $\gUPSILON=\left[\left(\gMc{1}-\gM\right)\cdots\left(\gMc{\gL{\gC}}-\gM\right)\right]$
  \State compute eigenvectors $\gOMEGA$ and corresponding eigenvalues $\gTHETA$ of $\gSIGMAt$ through SVD of $\gCHI$
  \State compute eigenvectors $\gXI$ of $\gTHETA^{\nicefrac{-1}{2}}\gOMEGA^\top\gSIGMAb\gOMEGA\gTHETA^{\nicefrac{-1}{2}}$ through SVD of $\gTHETA^{\nicefrac{-1}{2}}\gOMEGA^\top\gUPSILON$
  \State compute eigenvectors $\gPSI=\gOMEGA\gTHETA^{\nicefrac{-1}{2}}\gXI$
  \State compute eigenvalues $\gDELTA=\gPSI^\top\gSIGMAb\gPSI$
  \State return transform $\gPHI$ as eigenvectors in $\gPSI$ that correspond to the eigenvalues of at least $\nicefrac{1}{2}$ in $\gDELTA$
\end{algorithmic}
\end{algorithm}

\begin{algorithm}[tb]
\caption{LearnTransformationMatrixPCALDA$\left(\gL{\gG}\right)$}
\label{a2}
\begin{algorithmic}[1]
  \State split $\gL{\gG}=\left\{\left(\gGn{n},\gLAMBDAn{n}\right)\right\}_{n=1}^{\gL{\gN}}$ into classes $\left\{\gIc{c}\right\}_{c=1}^{\gL{\gC}}$ of $\gNc{c}=\left|\gIc{c}\right|$ samples
  \State compute $\gSIGMAb=\sum_{c=1}^{\gL{\gC}}\left(\gMc{c}-\gM\right)\left(\gMc{c}-\gM\right)^\top$
  \State compute $\gSIGMAw=\sum_{c=1}^{\gL{\gC}}\frac{1}{\gNc{c}}\sum_{n=1}^{\gNc{c}}\left(\gGnc{n}{c}-\gMc{c}\right)\left(\gGnc{n}{c}-\gMc{c}\right)^\top$
  \State compute eigenvectors $\gPHI_\mathrm{PCA}$ of $\gSIGMAt=\gSIGMAb+\gSIGMAw$ that correspond to $\gO{\gD}$ largest eigenvalues (we set $\gO{\gD}=\gL{\gC}$)
  \State compute eigenvectors $\gPHI_\mathrm{LDA}$ of $(\gPHI_\mathrm{PCA}^\top\gSIGMAw\gPHI_\mathrm{PCA})^{-1}(\gPHI_\mathrm{PCA}^\top\gSIGMAb\gPHI_\mathrm{PCA})$
  \State return transform $\gPHI=\gPHI_\mathrm{PCA}\gPHI_\mathrm{LDA}$
\end{algorithmic}
\end{algorithm}

\section{Evaluation}
\label{eval}

This paper provides an extended evaluation. In the following we describe the evaluation database and framework, setups for data separation into learning and evaluation parts, evaluation metrics, and results with discussion.

\subsection{Database and Framework}
\label{eval-data}

For evaluation purposes we have extracted a large number of gait samples from the MoCap database obtained from the CMU Graphics Lab~\cite{CMU03}, which is available under the Creative Commons license. It is a well-known and recognized database of structural human motion data and contains a~considerable number of gait sequences. Motions are recorded with an optical marker-based Vicon system. People wear a black jumpsuit with 41~markers taped on. The tracking space of \unit[30]{m$^2$} is surrounded by 12~cameras with a sampling rate of \unit[120]{Hz} at heights ranging from 2 to 4~meters above ground thereby creating a video surveillance environment. Motion videos are triangulated to get highly accurate 3D data in the form of relative body point coordinates (with respect to the root joint) in each video frame and are stored in the standard ASF/AMC data format. Each registered participant is assigned with their respective skeleton described in an ASF file. Motions in the AMC files store bone rotational data, which is interpreted as instructions about how the associated skeleton deforms over time.

These MoCap data, however, contain skeleton parameters pre-calibrated by the CMU staff. Skeletons are unique to each walker and even a trivial skeleton check could result in 100\% recognition. In order to fairly use the collected data, a prototypical skeleton is constructed and used to represent bodies of all subjects, shrouding the skeleton parameters. Assuming that all walking individuals are physically identical disables the skeleton check from being a potentially unfair classifier. Moreover, this is a skeleton-robust solution as all bone rotational data are linked to one specific skeleton. To obtain realistic parameters, it is calculated as the mean of all skeletons in the provided ASF files.

The raw data are in the form of bone rotations or, if combined with the prototypical skeleton, 3D joint coordinates. The bone rotational data are taken from the AMC files without any pre-processing. We calculate the joint coordinates using the bone rotational data and the prototypical skeleton. One cannot directly use raw values of joint coordinates, as they refer to absolute positions in the tracking space, and not all potential methods are invariant to a person's position or walk direction. To ensure such invariance, the center of the coordinate system is moved to the position of root joint $\gGAMMAjt{\mathrm{root}}{t}=[0,0,0]^\top$ for each time $t$ and the axes are adjusted to the walker's perspective: the X~axis is from right (negative) to left (positive), the Y~axis is from down (negative) to up (positive), and the Z~axis is from back (negative) to front (positive). In the AMC file structure notation it is achieved by setting the root translation and rotation to zero (\texttt{root 0 0 0 0 0 0}) in all frames of all motion sequences.

Since the general motion database contains all motion types, we extracted a number of sub-motions that represent gait cycles. First, an exemplary gait cycle was identified, and clean gait cycles were then filtered out using a threshold for their DTW distance on bone rotations in time. The distance threshold was explicitly set low enough so that even the least similar sub-motions still semantically represent gait cycles. Setting this threshold higher might also qualify sub-motions that do not resemble gait cycles anymore. Finally, subjects that contributed with less than 10~samples were excluded. The final database~\cite{WWW} has 54~walking subjects that performed 3,843~samples in total, resulting in an average of about 71~samples per subject.

As a contribution to reproducible research, we release our database and framework to improve the development, evaluation and comparison of methods for gait recognition from MoCap data. They are intended also for our fellow researchers and reviewers to reproduce the results of our experiments. Our recent paper~\cite{BS16c} provides a manual and comments on reproducing the experiments. With this manual, a reader should be able to reproduce the evaluation and to use the implementation in their own application. The source codes and data are available online at our research group web page~\cite{WWW} with link to our departmental Git repository.

The evaluation framework comprises \textbf{(i)}~source codes of the state-of-the-art human-interpretable geometric features that we have reviewed in our work to date as well as own our two approaches where gait features are learned by MMC (see Algorithm~\ref{a1}) and by PCA+LDA (see Algorithm~\ref{a2}), the Random method (no features and random classifier), and the Raw method (raw data and DTW distance function). Depending on whether the raw data are in the form of bone rotations or joint coordinates, the methods are referred to with BR or JC subscripts, respectively. The framework includes \textbf{(ii)}~a~mechanism for evaluating four class separability coefficients of feature space and four classifier performance metrics. It also contains \textbf{(iii)}~a~tool for learning a custom classifier and for classifying a custom probe on a custom gallery. We provide \textbf{(iv)}~an experimental database along with source codes for its extraction from the CMU MoCap database.

\subsection{Evaluation Setup and Metrics}
\label{eval-setup}

\begin{figure}[tb]
\centering\small
\def\blackbox{
\tikzstyle{important line}=[very thick]
\draw[style=important line] 
    (0,0) --node[left]{\rotatebox{90}{\hbox{Identities}}}
    (0,6) --node[above]{Samples}
    (9,6) -- (9,0) -- (0,0) -- cycle;}
\begin{tikzpicture}[scale=0.6]
\blackbox \node at (4.5,-.8) {Homogeneous setup};
\draw[fill=ACMBlue!40!white] (0,0) -- (3,0) -- (3,6) -- (0,6) -- cycle ;
\draw[fill=ACMGreen!40!white] (3,0) -- (9,0) -- (9,6) -- (3,6) -- cycle ;
\node at (6,3) {Evaluation};
\node at (1.5,3) {Learning};
\end{tikzpicture}
\qquad
\begin{tikzpicture}[scale=0.6]
\blackbox \node at (4.5,-.8) {Heterogeneous setup};
\draw[fill=ACMGreen!40!white] (0,0) -- (9,0) -- (9,4) -- (0,4) -- cycle ;
\draw[fill=ACMBlue!40!white] (0,4) -- (9,4) -- (9,6) -- (0,6) -- cycle ;
\node at (4.5,2) {Evaluation};
\node at (4.5,5) {Learning};
\end{tikzpicture}
\caption{Abstraction of the concept of homogeneous and heterogeneous setups. Consider a matrix (visualized as a rectangle) of all database samples where each row contains samples of one identity class. The homogeneous setup separates all identities into $\nicefrac{1}{3}$ learning samples and $\nicefrac{2}{3}$ evaluation samples (left), whereas the heterogeneous setup takes $\nicefrac{1}{3}$ learning identities and $\nicefrac{2}{3}$ evaluation identities with all their samples (right). Setting the sizes of learning and evaluation parts is a subject of setup configuration.}
\label{f2}
\end{figure}
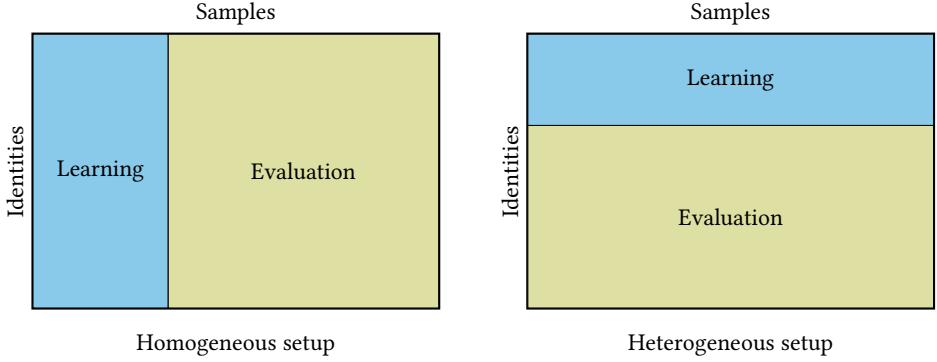

In the following, we introduce two setups of data separation: homogeneous and heterogeneous. The homogeneous setup learns the transformation matrix on a fraction of samples of $\gL{\gC}$ identities and is evaluated on templates derived from the remaining samples of the same $\gE{\gC}=\gL{\gC}$ identities. The heterogeneous setup learns the transform on all samples of $\gL{\gC}$ identities and is evaluated on all templates derived from all samples of other $\gE{\gC}$ identities. An abstraction of this concept is depicted in Figure~\ref{f2}. Note that in heterogeneous setup no walker identity is ever used for both learning and evaluation at the same time.

A use of the homogeneous setup can be any system to recognize people that cooperated at learning the features. For example, a small company entrance authentication where all employees cooperated to register. MoCap data of each employee's gait pattern can be registered along with a picture of their face. On the other hand, the heterogeneous setup can be used at person re-identification. During video surveillance, new identities can appear on the fly and labeled data for all the people encountered may not always be available. In this scenario, forensic investigators may ask for tracking the suspects.

The homogeneous setup is parametrized by a single number $\gL{\gC}=\gE{\gC}$ of learning-and-evaluation identity classes, whereas the heterogeneous setup has the form $\left(\gL{\gC},\gE{\gC}\right)$. This parametrization specifies how many learning and how many evaluation identity classes are randomly selected from the database. The evaluation of each setup is repeated 3~times, selecting new random identity classes each time and reporting the average result.

In both setups, the class separability coefficients are calculated directly on the full evaluation part whereas the classification metrics are estimated with the 10-fold cross-validation taking one dis-labeled fold as a testing set and the other nine labeled folds as gallery. Test templates are classified by the winner-takes-all strategy, in which a test template $\gGnH{}^{\mathrm{test}}$ gets assigned with the label $\gLAMBDAn{\argmin_i\gDELTAccH{\gGnH{}^{\mathrm{test}}}{\gGnH{i}^{\mathrm{gallery}}}}$ of the gallery's closest identity class.

What follows is a list of evaluation metrics. Recognition rate is often perceived as the ultimate qualitative measure; however, it is more explanatory to include an evaluation in terms of class separability of the feature space. The class separability measures give an estimate of the recognition potential of the extracted features and do not reflect eventual combination with an unsuitable classifier:
\begin{description}[style=unboxed,leftmargin=0cm]
\def\ditem #1 (#2){\item[\textbullet\enskip\normalfont{\textit{#1:}}\enskip\textbf{#2}]}
\ditem Davies-Bouldin Index (DBI)
\begin{equation}
\mathrm{DBI}=\frac{1}{\gE{\gC}}\sum_{c=1}^{\gE{\gC}}\max\limits_{1 \leq c' \leq \gE{\gC},\,c' \neq c}\frac{\sigma_c+\sigma_{c'}}{\gDELTAccH{\gMcH{c}}{\gMcH{c'}}}
\end{equation}
where $\sigma_c=\frac{1}{\gNc{c}}\sum_{n=1}^{\gNc{c}}\gDELTAccH{\gGnH{n}}{\gMcH{c}}$ is the average distance of all elements in identity class $\gIc{c}$ to its centroid, and analogically for $\sigma_{c'}$. Templates of low intra-class distances and of high inter-class distances have a low DBI.
\ditem Dunn Index (DI)
\begin{equation}
\mathrm{DI}=\frac{\min\limits_{1 \leq c<c' \leq \gE{\gC}}\gDELTAccH{\gMcH{c}}{\gMcH{c'}}}{\max\limits_{1 \leq c \leq \gE{\gC}}\sigma_c}
\end{equation}
with $\sigma_c$ from the above DBI. Since this criterion seeks classes with high intra-class similarity and low inter-class similarity, a high DI is more desirable.
\ditem Silhouette Coefficient (SC)
\begin{equation}
\mathrm{SC}=\frac{1}{\gE{\gN}}\sum_{n=1}^{\gE{\gN}}\frac{b\!\left(\gGnH{n}\right)-a\!\left(\gGnH{n}\right)}{\max\left\{a\!\left(\gGnH{n}\right),b\!\left(\gGnH{n}\right)\right\}}
\end{equation}
where $a\!\left(\gGnH{n}\right)=\frac{1}{\gNc{c}}\sum_{n'=1}^{\gNc{c}}\gDELTAccH{\gGnH{n}}{\gGnH{n'}}$ is the average distance from $\gGnH{n}$ to other samples within the same identity class and $b\!\left(\gGnH{n}\right)=\min\limits_{1 \leq c' \leq \gE{\gC},\,c' \neq c}\frac{1}{\gNc{c'}}\sum_{n'=1}^{\gNc{c'}}\gDELTAccH{\gGnH{n}}{\gGnH{n'}}$ is the average distance of $\,\gGnH{n}$ to the samples in the closest class. It is clear that $-1\leq\mathrm{SC}\leq1$ and a SC close to one mean that classes are appropriately separated.
\ditem Fisher's Discriminant Ratio (FDR)
\begin{equation}
\mathrm{FDR}=\frac{\frac{1}{\gE{\gC}}\sum_{c=1}^{\gE{\gC}}\gDELTAccH{\gMcH{c}}{\gH{\gM}}}{\frac{1}{\gE{\gN}}\sum_{c=1}^{\gE{\gC}}\sum_{n=1}^{\gNc{c}}\gDELTAccH{\gGnH{n}}{\gMcH{c}}}.
\end{equation}
A high FDR is preferred for seeking classes with low intra-class sparsity and high inter-class sparsity.
\end{description}

Apart from analyzing the distribution of templates in the feature space, it is schematic to combine the features with a rank-based classifier and to evaluate the system based on distance distribution with respect to a query. For obtaining a more applied performance evaluation, we evaluate:
\begin{description}[style=unboxed,leftmargin=0cm]
\def\ditem #1 (#2){\item[\textbullet\enskip\normalfont{\textit{#1:}}\enskip\textbf{#2}]~\newline}
\ditem Cumulative Match Characteristic (CMC) 
Sequence of Rank-$k$ (for $k$ on X axis from 1 up to $\gE{\gC}$) recognition rates (Y~axis) to measure ranking capabilities of a recognition method. Its headline Rank-1 is the Correct Classification Rate (\textbf{CCR}).
\ditem False Accept Rate vs.\@ False Reject Rate (FAR/FRR)
Two sequences of the error rates (Y~axis) as functions of the discrimination threshold (X~axis). Each method has a value $e$ of this threshold giving Equal Error Rate (\textbf{EER}=FAR=FRR).
\ditem Receiver Operating Characteristic (ROC)
Sequence of True Accept Rate (TAR) and False Accept Rate with a varied discrimination threshold. For a given threshold, the system signalizes both TAR (Y~axis) and FAR (X~axis). The value of Area Under Curve (\textbf{AUC}) is computed as the integral of the ROC curve.
\ditem Recall vs.\@ Precision (RCL/PCN)
Sequence of the rates with a varied discrimination threshold. For a given threshold the system signals both RCL (X~axis) and PCN (Y~axis). The value of Mean Average Precision (\textbf{MAP}) is computed as the area under the RCL/PCN curve.
\end{description}

These measures reflect how class-separated the feature space is, how often the walk pattern of a person is classified correctly and how difficult it is to confuse different people. They do not, in fact, provide complementary information, although a quality evaluation framework should be able to evaluate the most popular measures. Different applications use different evaluation measures. For example, a hotel lobby authentication system could use a high Rank-3 at the CMC, while a city-level person tracking system is likely to need the ROC curve leaning towards the upper left corner.

Finally, the evaluation incorporates two scalability measures: average distance computation time (\textbf{DCT}) in milliseconds and average template dimensionality~(\textbf{TD}).

\subsection{Results}
\label{eval-res}

In this section we provide comparative evaluation results of the feature extraction methods in terms of evaluation metrics defined in Section~\ref{eval-setup}. To ensure a fair comparison, we evaluate all methods on the same experimental database and framework described in Section~\ref{eval-data}. Table~\ref{t1} presents the implementation details of all the methods and the results of class separability coefficients, classification metrics and scalability.

\afterpage{
\begin{landscape}
\begin{table}[tb]
\caption{Class separability coefficients, classification metrics and scalability metrics for all implemented methods. Homogeneous setup, evaluated on the full database with nested cross-validation. We advise readers to read the original papers for better understanding of the implementation column. The last column contains the total evaluation times (seconds~\unit{s}, minutes~\unit{m}, hours~\unit{h}, days~\unit{d}).}
\label{t1}
\centering\tabcolsep3.2pt
\begin{tabular}{r|cc|rlll|llll|rr|l}
\toprule[1pt]
& \multicolumn{2}{c|}{implementation}	& \multicolumn{4}{c|}{class separability coefficients}	& \multicolumn{4}{c|}{classification metrics}	& \multicolumn{2}{c|}{scalability}	& evaluation \\
method	& gait features	& dist. function	& DBI\hphantom{0}	& \hphantom{0}DI	& \hphantom{0}SC	& FDR	& CCR	& EER	& AUC	& MAP	& DCT	& TD	& total time \\
\midrule[0.4pt]
AhmedF	& JRD + JRA	& DTW+$L_1$	& 185.8	& 1.708	& $-$0.164	& 0.77	& 0.891	& 0.374	& 0.677	& 0.254	& 45	& 5,254	& \unit[\hphantom{0}10.3]{d} \\
AhmedM	& HDF + VDF	& $L_1$	& 216.2	& 0.842	& $-$0.246	& 0.954	& 0.657	& 0.38	& 0.659	& 0.165	& \textbf{<1}	& 24	& \unit[\hphantom{0}48.6]{m} \\
AliS	& triangle areas	& $L_1$	& 501.5	& 0.26	& $-$0.463	& 1.175	& 0.225	& 0.384	& 0.679	& 0.111	& \textbf{<1}	& 2	& \unit[\hphantom{0}40.9]{m} \\
AnderssonVO	& all described	& $L_1$	& 142.3	& 1.297	& $-$0.102	& 1.127	& 0.84	& 0.343	& 0.715	& 0.251	& \textbf{<1}	& 68	& \unit[\hphantom{0}45.7]{m} \\
BallA	& all described	& $L_1$	& 161\hphantom{.0}	& 1.458	& $-$0.163	& 1.117	& 0.75	& 0.346	& 0.711	& 0.231	& \textbf{<1}	& 18	& \unit[\hphantom{0}48.5]{m} \\
DikovskiB	& Dataset 3	& $L_1$	& 144.5	& 1.817	& $-$0.135	& \textbf{1.227}	& 0.881	& 0.363	& 0.695	& 0.254	& \textbf{<1}	& 71	& \unit[\hphantom{0}50.7]{m} \\
JiangS	& angles	& DTW+$L_1$	& 206.6	& 1.802	& $-$0.249	& 0.85	& 0.811	& 0.395	& 0.657	& 0.242	& 8	& 584	& \unit[\hphantom{00}1.9]{d} \\
KrzeszowskiT	& all described	& DTW+$L_1$	& 154.1	& 1.982	& $-$0.147	& 0.874	& 0.915	& 0.392	& 0.662	& 0.275	& 35	& 3,795	& \unit[\hphantom{00}8.1]{d} \\
KwolekB	& \texttt{g\_all}	& $L_1$	& 150.9	& 1.348	& $-$0.084	& 1.175	& 0.896	& 0.358	& 0.723	& 0.323	& \textbf{<1}	& 660	& \unit[\hphantom{00}1.1]{h} \\
NareshKumarMS	& normalized JC	& cov. matrices	& \textbf{118.6}	& 1.618	& $-$0.086	& 1.09	& 0.801	& 0.459	& 0.631	& 0.217	& 8	& 13,950	& \unit[\hphantom{00}1.8]{d} \\
PreisJ	& static + dynamic	& $L_1$	& 1,980.6	& 0.055	& $-$0.512	& 1.067	& 0.143	& 0.401	& 0.626	& 0.067	& \textbf{<1}	& 13	& \unit[\hphantom{0}48.7]{m} \\
SedmidubskyJ	& $\mathcal{S}_{C_LH_L}+\mathcal{S}_{C_RH_R}$	& DTW+$L_1$	& 398.1	& 1.35	& $-$0.425	& 0.811	& 0.543	& 0.388	& 0.657	& 0.149	& \textbf{<1}	& 292	& \unit[\hphantom{00}1.4]{d} \\
SinhaA	& all described	& $L_1$	& 214.8	& 1.112	& $-$0.215	& 1.101	& 0.674	& 0.356	& 0.697	& 0.191	& \textbf{<1}	& 45	& \unit[\hphantom{0}49.6]{m} \\
\midrule[0.4pt]
MMC\sub{BR}	& MMC on BR	& Mahalanobis	& 154.2	& 1.638	& \textbf{\hphantom{$-$}0.062}	& 1.173	& 0.925	& \textbf{0.297}	& \textbf{0.748}	& 0.353	& \textbf{<1}	& 53	& \unit[\hphantom{00}2.6]{h} \\
MMC\sub{JC}	& MMC on JC	& Mahalanobis	& 130.3	& 1.891	& \hphantom{$-$}0.051	& 1.106	& 0.918	& 0.378	& 0.721	& 0.315	& \textbf{<1}	& 51	& \unit[\hphantom{00}3.0]{h} \\
PCA+LDA\sub{BR}	& PCA+LDA on BR	& Mahalanobis	& 182\hphantom{.0}	& 1.596	& $-$0.015	& 0.984	& 0.918	& 0.361	& 0.695	& 0.276	& \textbf{<1}	& 54	& \unit[\hphantom{00}4.7]{h} \\
PCA+LDA\sub{JC}	& PCA+LDA on JC	& Mahalanobis	& 174.4	& 1.309	& $-$0.091	& 0.827	& 0.863	& 0.44	& 0.643	& 0.201	& \textbf{<1}	& 54	& \unit[\hphantom{0}10.9]{h} \\
Random	& none	&	&	&	&	&	& 0.042	&	&	&	&	& \textbf{0}	& \unit[\hphantom{0}27.9]{m} \\
Raw\sub{BR}	& BR	& DTW+$L_2$	& 163.7	& \textbf{2.092}	& \hphantom{$-$}0.011	& 0.948	& \textbf{0.966}	& 0.315	& 0.743	& \textbf{0.358}	& 70	& 8,229	& \unit[\hphantom{0}16.1]{d}\\
Raw\sub{JC}	& JC	& DTW+$L_2$	& 155.1	& 1.954	& $-$0.12	& 0.897	& 0.926	& 0.377	& 0.679	& 0.283	& 161	& 13,574	& \unit[\hphantom{0}36.7]{d} \\
\bottomrule[1pt]
\end{tabular}
\end{table}
\end{landscape}
}

The goal of the MMC-based learning is to find a linear discriminant that maximizes the misclassification margin. This optimization technique appears to be more effective than designing geometric gait features. A variety of class-separability coefficients and classification metrics allows insights from different statistical perspectives. The results in Table~\ref{t1} indicate that the proposed MMC\sub{BR} method (on bone rotational data) is a leading concept for rank-based classifier systems: highest SC and AUC, lowest EER, and competitive DBI, DI, FDR, CCR and MAP. In terms of recognition rate, the MMC method was only outperformed by the Raw method, which is implemented here as a form of baseline. We interpret the high scores as a sign of robustness.

Apart from the performance merits, the MMC method is also efficient: relatively low-dimensional templates and Mahalanobis distance ensure fast distance computations and thus contribute to high scalability. Note that even if the Raw method has some of the best results, it can hardly be used in practice due to its extreme consumption of time and space resources. On the other hand, Random has no features but cannot be considered a serious recognition method. To illustrate the evaluation time, calculating the distance matrix (a matrix of distances between all evaluation templates) took a couple minutes for the MMC method, almost nothing for the Random method, and more than two weeks for the Raw method. The learning time of the MMC and PCA+LDA methods increases with the number of learning samples $\gL{\gN}$ (counting $\gSIGMAw$ and $\gSIGMAt$ matrices); however, this is not an issue in the field of gait recognition as training the models suffers from the opposite problem, undersampling.

To reproduce the experiments in Table~\ref{t1}, instructions to follow are to be found at~\cite{WWW}. Please note that some methods are rather slow~--~total evaluation times (learning included) in the last column were measured on a computer with Intel\textregistered\ Xeon\textregistered\ CPU E5-2650 v2 @ 2.60GHz and \unit[256]{GB}~RAM.

An additional experiment was carried out with the traditional 10-fold cross-validation in which gait cycles of a common walk sequence are always kept together in the same fold. This is to prevent a situation in which two consecutive gait cycles are split between testing and training folds which would cause a potential overtraining. The classification metrics of methods include CCR, EER, AUC, MAP, and the CMC curve up to Rank-10. Displayed in Table~\ref{t2}, the results suggest that the methods based on machine learning outperform the methods of hand-designed features in terms of all measured metrics. Note that the results are slightly worse than those in Table~\ref{t1} due to avoiding the overtraining cases and that the order of the tested methods is preserved. We interpret this order as a demonstrative result of comparison of all tested methods based on their classification potential.

\begin{table}[tb]
\caption{Classification metrics for all implemented methods. Homogeneous setup, evaluated on the full database with 10-fold cross-validation. Methods are ordered by their CCR score.}
\label{t2}
\centering\tabcolsep3pt
\begin{tabular}{r|llll|ll}
\toprule[1pt]
method	& CCR	& EER	& AUC	& MAP	& CMC \\
\midrule[0.4pt]
Raw\sub{JC}	& 0.872	& 0.321	& 0.731	& 0.317	& \multirow{1}{*}{\includegraphics[height=240pt]{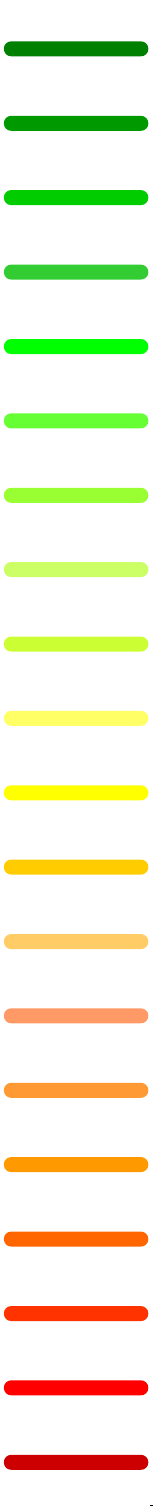}}	& \multirow{1}{*}{\includegraphics[height=240pt]{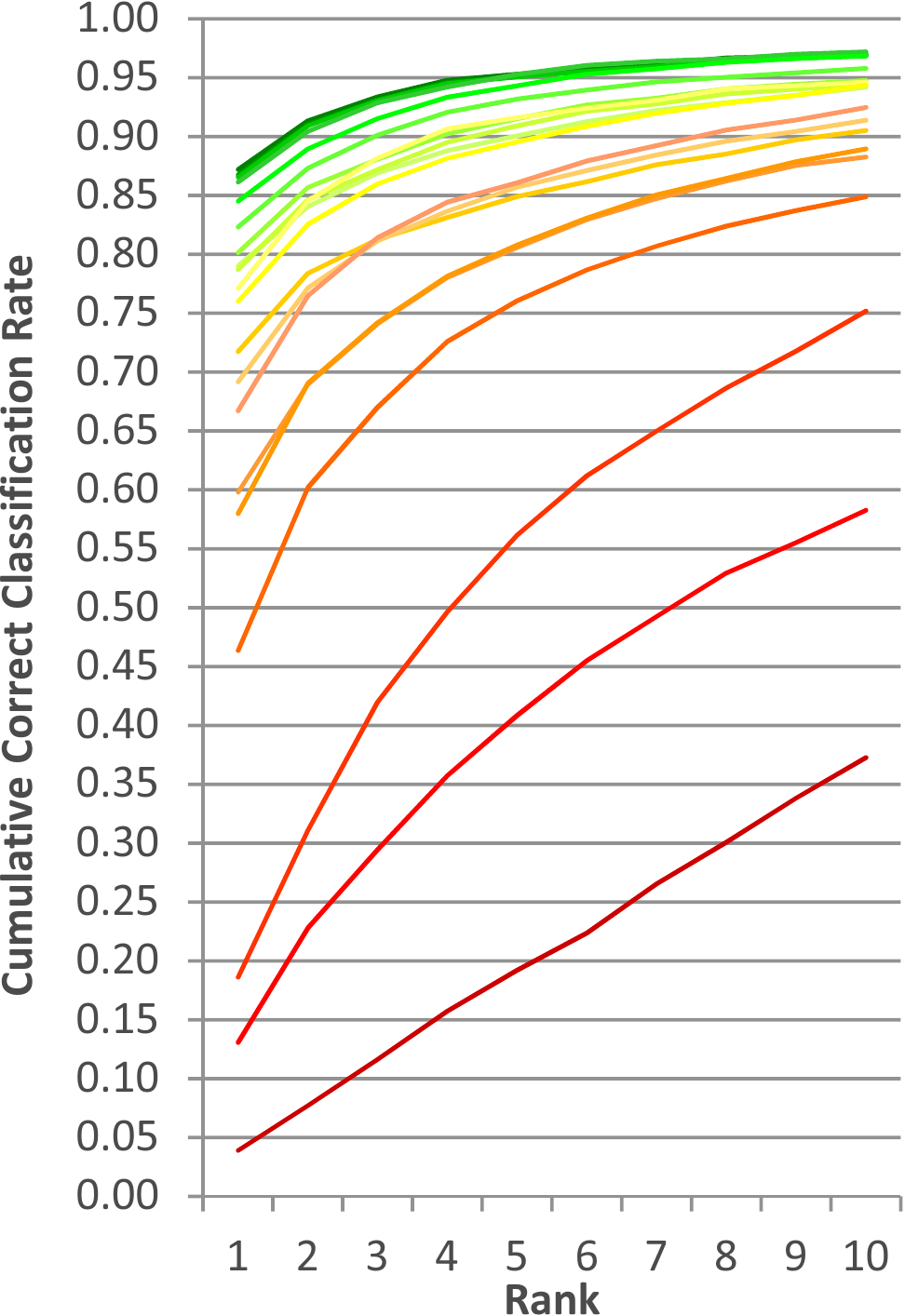}} \\
MMC\sub{BR}	& 0.868	& 0.305	& 0.739	& 0.332 \\
Raw\sub{BR}	& 0.867	& 0.333	& 0.701	& 0.259 \\
MMC\sub{JC}	& 0.861	& 0.325	& 0.72	& 0.309 \\
PCA+LDA\sub{BR}	& 0.845	& 0.335	& 0.682	& 0.247 \\
KwolekB	& 0.823	& 0.367	& 0.711	& 0.296 \\
KrzeszowskiT	& 0.802	& 0.348	& 0.717	& 0.273 \\
PCA+LDA\sub{JC}	& 0.79	& 0.417	& 0.634	& 0.189 \\
DikovskiB	& 0.787	& 0.376	& 0.679	& 0.227 \\
AhmedF	& 0.771	& 0.371	& 0.664	& 0.22 \\
AnderssonVO	& 0.76	& 0.352	& 0.703	& 0.228 \\
NareshKumarMS	& 0.717	& 0.459	& 0.613	& 0.19 \\
JiangS	& 0.692	& 0.407	& 0.637	& 0.204 \\
BallA	& 0.667	& 0.356	& 0.698	& 0.207 \\
SinhaA	& 0.598	& 0.362	& 0.69	& 0.176 \\
AhmedM	& 0.58	& 0.392	& 0.646	& 0.145 \\
SedmidubskyJ	& 0.464	& 0.394	& 0.65	& 0.138 \\
AliS	& 0.186	& 0.394	& 0.662	& 0.096 \\
PreisJ	& 0.131	& 0.407	& 0.618	& 0.066 \\
Random	& 0.039	& & & \\
\bottomrule[1pt]
\end{tabular}
\end{table}

\begin{figure}[tb]
\raggedright
\includegraphics[height=40mm]{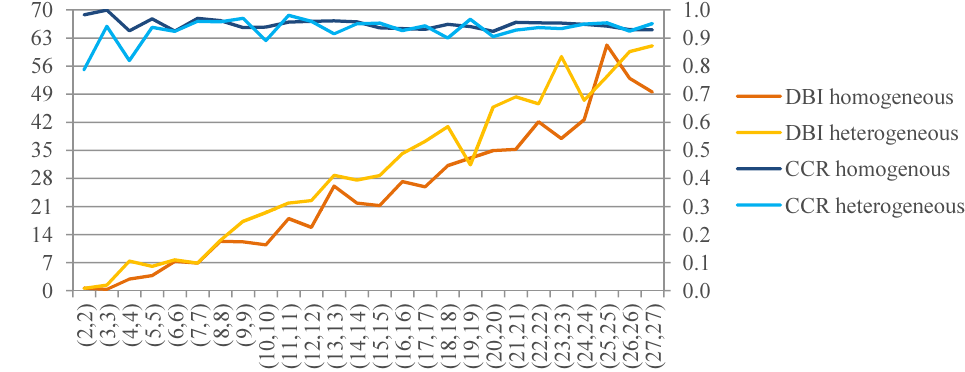}\\[\smallskipamount]
\includegraphics[height=40mm]{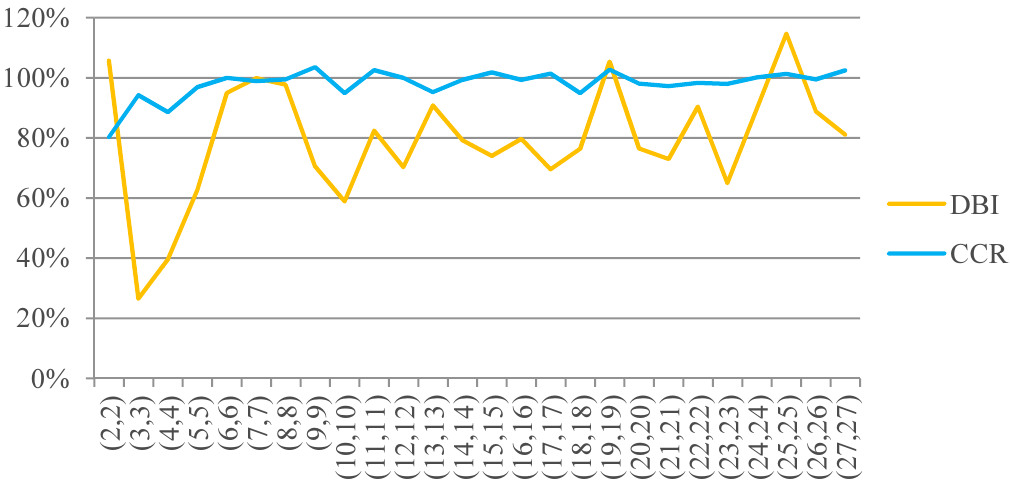}
\caption{DBI (left vertical axis) and CCR (right vertical axis) for experiments~\exper{A} of homogeneous setup and~\exper{B} of heterogeneous setup (top) with $\left(\gL{\gC},\gE{\gC}\right)$ configurations (horizontal axes) and their percentages (bottom).}
\label{f3}
\end{figure}

\begin{figure}[tb]
\raggedright
\includegraphics[height=40mm]{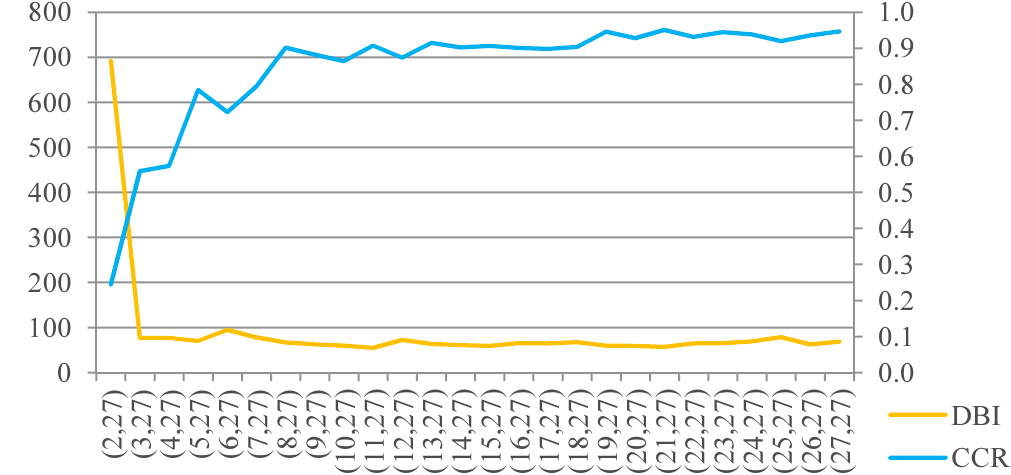}\\[\smallskipamount]
\includegraphics[height=40mm]{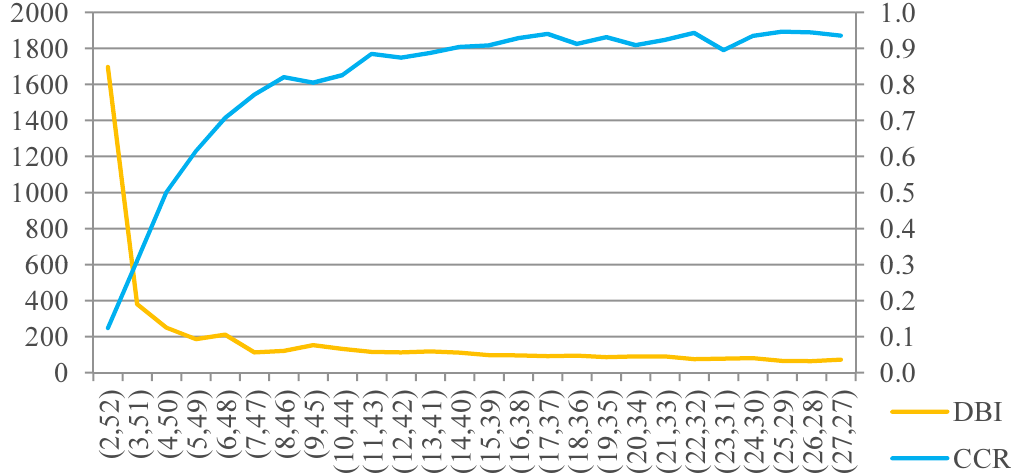}
\caption{DBI (left vertical axes) and CCR (right vertical axes) for experiments \exper{C} (top) and \exper{D} (bottom) on heterogeneous setup with $\left(\gL{\gC},\gE{\gC}\right)$ configurations (horizontal axes).}
\label{f4}
\end{figure}

Finally, the experiments \exper{A}, \exper{B}, \exper{C}, \exper{D} described below compare the homogeneous and heterogeneous setup and examine how quality of the system in the heterogeneous setup improves with an increased number of learning identities. The main idea is to show that we can learn what aspects of walk people generally differ in and extract those as general gait features. Recognizing people without needing group-specific features is convenient as particular people might not always provide annotated learning data. Results are illustrated in Figure~\ref{f3} and in Figure~\ref{f4}.
\begin{description}
\itemsep0pt
\parskip0pt
\item[\exper{A}] homogeneous setup with $\gL{\gC}=\gE{\gC}\in\left\{2,\ldots,27\right\}$;
\item[\exper{B}] heterogeneous setup with $\gL{\gC}=\gE{\gC}\in\left\{2,\ldots,27\right\}$;
\item[\exper{C}] heterogeneous setup with $\gL{\gC}\in\left\{2,\ldots,27\right\}$ and $\gE{\gC}=27$;
\item[\exper{D}] heterogeneous setup with $\gL{\gC}\in\left\{2,\ldots,27\right\}$ and $\gE{\gC}=27-\gL{\gC}$.
\end{description}

Experiments \exper{A} and \exper{B} compare homogeneous and heterogeneous setups by measuring the drop in the quality measures on an identical number of learning and evaluation identities ($\gL{\gC}=\gE{\gC}$). Please note that based on Section~\ref{eval-data} our database has 54~identity classes in total and they can be split into the learning and evaluation parts in the numbers of at most $\left(27,27\right)$. Top plot in Figure~\ref{f3} shows the values of DBI and CCR metrics in both alternatives, which not only appear comparable but also in some configurations the heterogeneous setup has an even higher CCR. Bottom plot expresses heterogeneous setup as a percentage of the homogeneous setup in each of the particular metrics. Here we see that with raising number of identities the heterogeneous setup approaches 100\% of the homogeneous setup. Therefore, the gait features learned by the introduced MMC method are walker-independent, which means that they can be learned on anybody.

Experiments~\exper{C} and~\exper{D} investigate the impact of the number of learning identities in the heterogeneous setup. Experiment~\exper{D} has only been evaluated up until $\left(27,27\right)$ as results with a learning part larger than an evaluation part would be insignificant. As can be seen in Figure~\ref{f4}, the performance grows quickly on the first configurations with very few learning identities, which we can interpret as an analogy to the Pareto (80--20) principle. Specifically, the results of experiment~\exper{C} say that 8~learning identities achieve almost the same performance (66.78~DBI and 0.902~CCR) as if they had been learned on 27~identities (68.32~DBI and 0.947~CCR). The outcome of experiment~\exper{D} indicates a similar growth of performance and we see that 14~identities can be enough to learn the transformation matrix to distinguish 40~completely different people (0.904~CCR).

\section{Conclusion}
\label{concl}

A common practice in state-of-the-art MoCap-based human identification is designing geometric gait features by hand. However, the field of pattern recognition has not recently advanced to a point where best results are frequently obtained using a machine learning approach. Finding optimal features for MoCap-based gait recognition is no exception. This work introduces the concept of learning robust and discriminative features directly from raw MoCap data by a modification of the Fisher Linear Discriminant Analysis with Maximum Margin Criterion and by combining Principal Component Analysis and Linear Discriminant Analysis, both with the goal of maximal separation of identity classes, avoiding the potential singularity problem caused by undersampling. Instead of instinctively drawing ad-hoc features, these methods are computed from a much larger space beyond the limits of human interpretability. The collection of features learned by MMC achieves leading scores in four class separability coefficients and therefore has a great potential for gait recognition applications. This is demonstrated on our extracted and normalized MoCap database of 54~walkers with 3,843~gait cycles by outperforming other 13~methods in numerous evaluation metrics. MMC is suitable for optimizing gait features; however, our future work will continue with research on further potential optimality criteria and multi-linear machine learning approaches.

The second contribution lies in showing the possibility of building a representation on a problem and using it on another (related) problem. Simulations on the CMU MoCap database show that our approach is able to build robust feature spaces without pre-registering and labeling all potential walkers. In fact, we can take different people (experiments~\exper{A} and~\exper{B}) and use just a fraction of them (experiments~\exper{C} and~\exper{D}). We have observed that already on 5~identities the heterogeneous evaluation setup exceeds 95\% of the homogeneous setup and improves with an increasing volume of identities. This means that one does not have to rely on the availability of all walkers for learning; instead, the features can be learned on anybody. The CCR of over 90\% when learning on 14~identities and evaluating on other 40~identities shows that an MMC-based system once learned on a fixed set of walker identities can be used to distinguish many more additional walkers. This is particularly important for a system that supports video surveillance applications where the encountered walkers never supply labeled data. Multiple occurrences of individual walkers can now be linked together even without knowing their actual identities.

In the name of reproducible research, our third contribution is the provision of the evaluation framework and database. All our data and source codes have been made available~\cite{WWW} under the Creative Commons Attribution license (CC-BY) for database and the Apache~2.0 license for software, which grant free use and allow for experimental evaluation. We hope that they will contribute to a~smooth development and evaluation of further novel MoCap-based gait recognition methods. We encourage all readers and developers of MoCap-based gait recognition methods to contribute to the framework with new algorithms, data and improvements.

\bibliographystyle{ACM-Reference-Format}
\bibliography{ref}


\begin{thebibliography}{00}


\ifx \showCODEN    \undefined \def \showCODEN     #1{\unskip}     \fi
\ifx \showDOI      \undefined \def \showDOI       #1{#1}\fi
\ifx \showISBNx    \undefined \def \showISBNx     #1{\unskip}     \fi
\ifx \showISBNxiii \undefined \def \showISBNxiii  #1{\unskip}     \fi
\ifx \showISSN     \undefined \def \showISSN      #1{\unskip}     \fi
\ifx \showLCCN     \undefined \def \showLCCN      #1{\unskip}     \fi
\ifx \shownote     \undefined \def \shownote      #1{#1}          \fi
\ifx \showarticletitle \undefined \def \showarticletitle #1{#1}   \fi
\ifx \showURL      \undefined \def \showURL       {\relax}        \fi
\providecommand\bibfield[2]{#2}
\providecommand\bibinfo[2]{#2}
\providecommand\natexlab[1]{#1}
\providecommand\showeprint[2][]{arXiv:#2}

\bibitem[\protect\citeauthoryear{Ahmed, Paul, and Gavrilova}{Ahmed
  et~al\mbox{.}}{2015}]%
        {APG15}
\bibfield{author}{\bibinfo{person}{Faisal Ahmed}, \bibinfo{person}{Padma~Polash
  Paul}, {and} \bibinfo{person}{Marina~L. Gavrilova}.}
  \bibinfo{year}{2015}\natexlab{}.
\newblock \showarticletitle{{DTW-Based Kernel and Rank-Level Fusion for 3D Gait
  Recognition Using Kinect}}.
\newblock \bibinfo{journal}{{\em The Visual Computer\/}} \bibinfo{volume}{31},
  \bibinfo{number}{6} (\bibinfo{year}{2015}), \bibinfo{pages}{915--924}.
\newblock
\showISSN{1432-2315}
\showDOI{%
\url{https://doi.org/10.1007/s00371-015-1092-0}}


\bibitem[\protect\citeauthoryear{Ahmed, Al-Jawad, and Sabir}{Ahmed
  et~al\mbox{.}}{2014}]%
        {AAS14}
\bibfield{author}{\bibinfo{person}{Mohammed Ahmed}, \bibinfo{person}{Naseer
  Al-Jawad}, {and} \bibinfo{person}{Azhin Sabir}.}
  \bibinfo{year}{2014}\natexlab{}.
\newblock \showarticletitle{{Gait Recognition Based on Kinect Sensor}}.
\newblock \bibinfo{journal}{{\em Proc. SPIE\/}}  \bibinfo{volume}{9139}
  (\bibinfo{year}{2014}), \bibinfo{pages}{91390B--91390B--10}.
\newblock
\showDOI{%
\url{https://doi.org/10.1117/12.2052588}}


\bibitem[\protect\citeauthoryear{Ali, Wu, Li, Saeed, Wang, and Zhou}{Ali
  et~al\mbox{.}}{2016}]%
        {AWLSWZ16}
\bibfield{author}{\bibinfo{person}{Sajid Ali}, \bibinfo{person}{Zhongke Wu},
  \bibinfo{person}{Xulong Li}, \bibinfo{person}{Nighat Saeed},
  \bibinfo{person}{Dong Wang}, {and} \bibinfo{person}{Mingquan Zhou}.}
  \bibinfo{year}{2016}\natexlab{}.
\newblock \bibinfo{booktitle}{{\em {Transactions on Computational Science XXVI:
  Special Issue on Cyberworlds and Cybersecurity}}}.
\newblock \bibinfo{publisher}{Springer}, \bibinfo{address}{Berlin, Heidelberg},
  Chapter {Applying Geometric Function on Sensors 3D Gait Data for Human
  Identification}, \bibinfo{pages}{125--141}.
\newblock
\showISBNx{978-3-662-49247-5}
\showDOI{%
\url{https://doi.org/10.1007/978-3-662-49247-5_8}}


\bibitem[\protect\citeauthoryear{Andersson and Araujo}{Andersson and
  Araujo}{2015}]%
        {AA15}
\bibfield{author}{\bibinfo{person}{Virginia~O. Andersson} {and}
  \bibinfo{person}{Ricardo~M. Araujo}.} \bibinfo{year}{2015}\natexlab{}.
\newblock \showarticletitle{{Person Identification Using Anthropometric and
  Gait Data from Kinect Sensor}}. In \bibinfo{booktitle}{{\em {Proc. of the
  Twenty-Ninth AAAI Conference on Artificial Intelligence (AAAI-15)}}}.
  \bibinfo{publisher}{AAAI Press}, \bibinfo{pages}{425--431}.
\newblock
\showURL{%
\url{http://www.aaai.org/ocs/index.php/AAAI/AAAI15/paper/view/9680}}


\bibitem[\protect\citeauthoryear{Balazia and Sojka}{Balazia and Sojka}{2016a}]%
        {BS16a}
\bibfield{author}{\bibinfo{person}{Michal Balazia} {and} \bibinfo{person}{Petr
  Sojka}.} \bibinfo{year}{2016}\natexlab{a}.
\newblock \showarticletitle{{Learning Robust Features for Gait Recognition by
  Maximum Margin Criterion}}. In \bibinfo{booktitle}{{\em {Proc. of 23rd
  International Conference on Pattern Recognition, ICPR 2016}}}.
  \bibinfo{publisher}{IEEE}, \bibinfo{pages}{901--906}.
\newblock
\showDOI{%
\url{https://doi.org/10.1109/ICPR.2016.7899750}}


\bibitem[\protect\citeauthoryear{Balazia and Sojka}{Balazia and Sojka}{2016b}]%
        {BS16b}
\bibfield{author}{\bibinfo{person}{Michal Balazia} {and} \bibinfo{person}{Petr
  Sojka}.} \bibinfo{year}{2016}\natexlab{b}.
\newblock \showarticletitle{Walker-Independent Features for Gait Recognition
  from Motion Capture Data}. In \bibinfo{booktitle}{{\em Structural, Syntactic,
  and Statistical Pattern Recognition: Joint IAPR International Workshop,
  S+SSPR 2016, M{\'e}rida, Mexico, November 29--December 2, 2016,
  Proceedings}}, \bibfield{editor}{\bibinfo{person}{Antonio Robles-Kelly},
  \bibinfo{person}{Marco Loog}, \bibinfo{person}{Battista Biggio},
  \bibinfo{person}{Francisco Escolano}, {and} \bibinfo{person}{Richard Wilson}}
  (Eds.). \bibinfo{publisher}{Springer International Publishing},
  \bibinfo{address}{Cham}, \bibinfo{pages}{310--321}.
\newblock
\showISBNx{978-3-319-49055-7}
\showDOI{%
\url{https://doi.org/10.1007/978-3-319-49055-7_28}}


\bibitem[\protect\citeauthoryear{Balazia and Sojka}{Balazia and Sojka}{2017a}]%
        {BS16c}
\bibfield{author}{\bibinfo{person}{Michal Balazia} {and} \bibinfo{person}{Petr
  Sojka}.} \bibinfo{year}{2017}\natexlab{a}.
\newblock \showarticletitle{An Evaluation Framework and Database for
  MoCap-Based Gait Recognition Methods}. In \bibinfo{booktitle}{{\em
  Reproducible Research in Pattern Recognition: First International Workshop,
  RRPR 2016, Canc{\'u}n, Mexico, December 4, 2016, Revised Selected Papers}},
  \bibfield{editor}{\bibinfo{person}{Bertrand Kerautret},
  \bibinfo{person}{Miguel Colom}, {and} \bibinfo{person}{Pascal Monasse}}
  (Eds.). \bibinfo{publisher}{Springer International Publishing},
  \bibinfo{address}{Cham}, \bibinfo{pages}{33--47}.
\newblock
\showISBNx{978-3-319-56414-2}
\showDOI{%
\url{https://doi.org/10.1007/978-3-319-56414-2_3}}


\bibitem[\protect\citeauthoryear{Balazia and Sojka}{Balazia and Sojka}{2017b}]%
        {WWW}
\bibfield{author}{\bibinfo{person}{Michal Balazia} {and} \bibinfo{person}{Petr
  Sojka}.} \bibinfo{year}{2017}\natexlab{b}.
\newblock \bibinfo{title}{{Gait Recognition from Motion Capture Data}}.
\newblock   (\bibinfo{date}{Feb.} \bibinfo{year}{2017}).
\newblock
\showURL{%
Retrieved May 30, 2017 from \url{https://gait.fi.muni.cz/}}
\newblock
\shownote{{Faculty of Informatics, Masaryk University, Brno}.}


\bibitem[\protect\citeauthoryear{Ball, Rye, Ramos, and Velonaki}{Ball
  et~al\mbox{.}}{2012}]%
        {BRRV12}
\bibfield{author}{\bibinfo{person}{Adrian Ball}, \bibinfo{person}{David Rye},
  \bibinfo{person}{Fabio Ramos}, {and} \bibinfo{person}{Mari Velonaki}.}
  \bibinfo{year}{2012}\natexlab{}.
\newblock \showarticletitle{Unsupervised Clustering of People from 'Skeleton'
  Data}. In \bibinfo{booktitle}{{\em Proceedings of the Seventh Annual ACM/IEEE
  International Conference on Human-Robot Interaction}} {\em
  (\bibinfo{series}{HRI '12})}. \bibinfo{publisher}{ACM}, \bibinfo{address}{New
  York, NY, USA}, \bibinfo{pages}{225--226}.
\newblock
\showISBNx{978-1-4503-1063-5}
\showDOI{%
\url{https://doi.org/10.1145/2157689.2157767}}


\bibitem[\protect\citeauthoryear{Belhumeur, Hespanha, and Kriegman}{Belhumeur
  et~al\mbox{.}}{1997}]%
        {BHK97}
\bibfield{author}{\bibinfo{person}{Peter~N. Belhumeur},
  \bibinfo{person}{Jo{\~a}o~P. Hespanha}, {and} \bibinfo{person}{David~J.
  Kriegman}.} \bibinfo{year}{1997}\natexlab{}.
\newblock \showarticletitle{{Eigenfaces vs.\@ Fisherfaces: Recognition Using
  Class Specific Linear Projection}}.
\newblock \bibinfo{journal}{{\em IEEE Trans. Pattern Anal. Mach. Intell.\/}}
  \bibinfo{volume}{19}, \bibinfo{number}{7} (\bibinfo{date}{July}
  \bibinfo{year}{1997}), \bibinfo{pages}{711--720}.
\newblock
\showISSN{0162-8828}
\showDOI{%
\url{https://doi.org/10.1109/34.598228}}


\bibitem[\protect\citeauthoryear{Boulgouris, Plataniotis, and
  Micheli-Tzanakou}{Boulgouris et~al\mbox{.}}{2009}]%
        {BPM09}
\bibfield{author}{\bibinfo{person}{N.~V. Boulgouris},
  \bibinfo{person}{Konstantinos~N. Plataniotis}, {and}
  \bibinfo{person}{Evangelia Micheli-Tzanakou}.}
  \bibinfo{year}{2009}\natexlab{}.
\newblock \bibinfo{booktitle}{{\em Biometrics: Theory, Methods, and
  Applications}}.
\newblock \bibinfo{publisher}{Wiley-IEEE Press}.
\newblock
\showISBNx{0470247827, 9780470247822}


\bibitem[\protect\citeauthoryear{Castro, Mar{\'{\i}}n{-}Jim{\'{e}}nez, Guil,
  and de~la Blanca}{Castro et~al\mbox{.}}{2016}]%
        {CMGP16}
\bibfield{author}{\bibinfo{person}{Francisco~M. Castro},
  \bibinfo{person}{Manuel~J. Mar{\'{\i}}n{-}Jim{\'{e}}nez},
  \bibinfo{person}{Nicol{\'{a}}s Guil}, {and}
  \bibinfo{person}{Nicolas~P{\'{e}}rez de~la Blanca}.}
  \bibinfo{year}{2016}\natexlab{}.
\newblock \showarticletitle{{Automatic Learning of Gait Signatures for People
  Identification}}.
\newblock \bibinfo{journal}{{\em CoRR\/}}  \bibinfo{volume}{abs/1603.01006}
  (\bibinfo{year}{2016}).
\newblock
\showURL{%
\url{https://arxiv.org/abs/1603.01006}}


\bibitem[\protect\citeauthoryear{Chaaraoui, Padilla-L{\'o}pez, and
  Fl{\'o}rez-Revuelta}{Chaaraoui et~al\mbox{.}}{2015}]%
        {CPF15}
\bibfield{author}{\bibinfo{person}{Alexandros~Andre Chaaraoui},
  \bibinfo{person}{Jos{\'e}~Ram{\'o}n Padilla-L{\'o}pez}, {and}
  \bibinfo{person}{Francisco Fl{\'o}rez-Revuelta}.}
  \bibinfo{year}{2015}\natexlab{}.
\newblock \showarticletitle{{Abnormal Gait Detection with RGB-D Devices using
  Joint Motion History Features}}. In \bibinfo{booktitle}{{\em 11th IEEE
  International Conference and Workshops on Automatic Face and Gesture
  Recognition (FG), 2015}}, Vol.~\bibinfo{volume}{7}. IEEE,
  \bibinfo{pages}{1--6}.
\newblock


\bibitem[\protect\citeauthoryear{Chen and Koskela}{Chen and Koskela}{2013}]%
        {CK13}
\bibfield{author}{\bibinfo{person}{Xi Chen} {and} \bibinfo{person}{Markus
  Koskela}.} \bibinfo{year}{2013}\natexlab{}.
\newblock \showarticletitle{{Classification of RGB-D and Motion Capture
  Sequences Using Extreme Learning Machine}}. In \bibinfo{booktitle}{{\em Proc.
  of Image Analysis: 18th Scandinavian Conference, SCIA 2013, Espoo, Finland,
  June 17--20, 2013}}, \bibfield{editor}{\bibinfo{person}{Joni-Kristian
  K{\"a}m{\"a}r{\"a}inen} {and} \bibinfo{person}{Markus Koskela}} (Eds.).
  \bibinfo{publisher}{Springer}, \bibinfo{pages}{640--651}.
\newblock
\showISBNx{978-3-642-38886-6}
\showDOI{%
\url{https://doi.org/10.1007/978-3-642-38886-6_60}}


\bibitem[\protect\citeauthoryear{Choudhury and Tjahjadi}{Choudhury and
  Tjahjadi}{2015}]%
        {CT15}
\bibfield{author}{\bibinfo{person}{Sruti~Das Choudhury} {and}
  \bibinfo{person}{Tardi Tjahjadi}.} \bibinfo{year}{2015}\natexlab{}.
\newblock \showarticletitle{{Robust View-Invariant Multiscale Gait
  Recognition}}.
\newblock \bibinfo{journal}{{\em Pattern Recognition\/}} \bibinfo{volume}{48},
  \bibinfo{number}{3} (\bibinfo{year}{2015}), \bibinfo{pages}{798--811}.
\newblock
\showISSN{0031-3203}
\showDOI{%
\url{https://doi.org/10.1016/j.patcog.2014.09.022}}


\bibitem[\protect\citeauthoryear{{CMU Graphics Lab}}{{CMU Graphics
  Lab}}{2003}]%
        {CMU03}
\bibfield{author}{\bibinfo{person}{{CMU Graphics Lab}}.}
  \bibinfo{year}{2003}\natexlab{}.
\newblock \bibinfo{title}{{Carnegie-Mellon Motion Capture (MoCap) Database}}.
\newblock   (\bibinfo{year}{2003}).
\newblock
\newblock
\shownote{\url{http://mocap.cs.cmu.edu}.}


\bibitem[\protect\citeauthoryear{Cutting and Kozlowski}{Cutting and
  Kozlowski}{1977}]%
        {CK77}
\bibfield{author}{\bibinfo{person}{James~E. Cutting} {and}
  \bibinfo{person}{Lynn~T. Kozlowski}.} \bibinfo{year}{1977}\natexlab{}.
\newblock \showarticletitle{Recognizing friends by their walk: Gait perception
  without familiarity cues}.
\newblock \bibinfo{journal}{{\em Bulletin of the Psychonomic Society\/}}
  \bibinfo{volume}{9}, \bibinfo{number}{5} (\bibinfo{year}{1977}),
  \bibinfo{pages}{353--356}.
\newblock
\showISSN{0090-5054}
\showDOI{%
\url{https://doi.org/10.3758/BF03337021}}


\bibitem[\protect\citeauthoryear{Devanne, Wannous, Daoudi, Berretti, Bimbo, and
  Pala}{Devanne et~al\mbox{.}}{2016}]%
        {DWDBDP16}
\bibfield{author}{\bibinfo{person}{Maxime Devanne}, \bibinfo{person}{Hazem
  Wannous}, \bibinfo{person}{Mohamed Daoudi}, \bibinfo{person}{Stefano
  Berretti}, \bibinfo{person}{Alberto~Del Bimbo}, {and} \bibinfo{person}{Pietro
  Pala}.} \bibinfo{year}{2016}\natexlab{}.
\newblock \showarticletitle{{Learning Shape Variations of Motion Trajectories
  for Gait Analysis}}. In \bibinfo{booktitle}{{\em Proc. of 23rd International
  Conference on Pattern Recognition, ICPR 2016, Canc\'un, Mexico}}.
  \bibinfo{pages}{895--900}.
\newblock
\showDOI{%
\url{https://doi.org/10.1109/ICPR.2016.7899749}}


\bibitem[\protect\citeauthoryear{Dikovski, Madjarov, and Gjorgjevikj}{Dikovski
  et~al\mbox{.}}{2014}]%
        {DMG14}
\bibfield{author}{\bibinfo{person}{Bojan Dikovski}, \bibinfo{person}{Gjorgji
  Madjarov}, {and} \bibinfo{person}{Dejan Gjorgjevikj}.}
  \bibinfo{year}{2014}\natexlab{}.
\newblock \showarticletitle{{Evaluation of Different Feature Sets for Gait
  Recognition Using Skeletal Data from Kinect}}. In \bibinfo{booktitle}{{\em
  37th Intl.\ Convention on Information and Communication Technology,
  Electronics and Microelectronics}}. \bibinfo{pages}{1304--1308}.
\newblock
\showDOI{%
\url{https://doi.org/10.1109/MIPRO.2014.6859769}}


\bibitem[\protect\citeauthoryear{Fisher}{Fisher}{1936}]%
        {F36}
\bibfield{author}{\bibinfo{person}{Ronald~A. Fisher}.}
  \bibinfo{year}{1936}\natexlab{}.
\newblock \showarticletitle{{The Use of Multiple Measurements in Taxonomic
  Problems}}.
\newblock \bibinfo{journal}{{\em Annals of Eugenics\/}} \bibinfo{volume}{7},
  \bibinfo{number}{2} (\bibinfo{year}{1936}), \bibinfo{pages}{179--188}.
\newblock
\showISSN{2050-1439}
\showDOI{%
\url{https://doi.org/10.1111/j.1469-1809.1936.tb02137.x}}


\bibitem[\protect\citeauthoryear{Han, Achar, Lee, and Pe{\~{n}}a-Mora}{Han
  et~al\mbox{.}}{2013}]%
        {HALP13}
\bibfield{author}{\bibinfo{person}{SangUk Han}, \bibinfo{person}{Madhav Achar},
  \bibinfo{person}{SangHyun Lee}, {and} \bibinfo{person}{Feniosky
  Pe{\~{n}}a-Mora}.} \bibinfo{year}{2013}\natexlab{}.
\newblock \showarticletitle{Empirical assessment of a RGB-D sensor on motion
  capture and action recognition for construction worker monitoring}.
\newblock \bibinfo{journal}{{\em Visualization in Engineering\/}}
  \bibinfo{volume}{1}, \bibinfo{number}{1} (\bibinfo{year}{2013}),
  \bibinfo{pages}{6}.
\newblock
\showISSN{2213-7459}
\showDOI{%
\url{https://doi.org/10.1186/2213-7459-1-6}}


\bibitem[\protect\citeauthoryear{Hong, Kang, and Price}{Hong
  et~al\mbox{.}}{2014}]%
        {HKP14}
\bibfield{author}{\bibinfo{person}{Jie Hong}, \bibinfo{person}{Jinsheng Kang},
  {and} \bibinfo{person}{Michael~E. Price}.} \bibinfo{year}{2014}\natexlab{}.
\newblock \showarticletitle{Extraction of bodily features for gait recognition
  and gait attractiveness evaluation}.
\newblock \bibinfo{journal}{{\em Multimedia Tools and Applications\/}}
  \bibinfo{volume}{71}, \bibinfo{number}{3} (\bibinfo{year}{2014}),
  \bibinfo{pages}{1999--2013}.
\newblock
\showISSN{1573-7721}
\showDOI{%
\url{https://doi.org/10.1007/s11042-012-1319-2}}


\bibitem[\protect\citeauthoryear{Jiang, Wang, Zhang, and Sun}{Jiang
  et~al\mbox{.}}{2015}]%
        {JWZS15}
\bibfield{author}{\bibinfo{person}{Shuming Jiang}, \bibinfo{person}{Yufei
  Wang}, \bibinfo{person}{Yuanyuan Zhang}, {and} \bibinfo{person}{Jiande Sun}.}
  \bibinfo{year}{2015}\natexlab{}.
\newblock \showarticletitle{{Real Time Gait Recognition System Based on Kinect
  Skeleton Feature}}.
\newblock In \bibinfo{booktitle}{{\em Computer Vision -- ACCV 2014 Workshops}},
  \bibfield{editor}{\bibinfo{person}{C.V. Jawahar} {and}
  \bibinfo{person}{Shiguang Shan}} (Eds.). \bibinfo{series}{LNCS},
  Vol.~\bibinfo{volume}{9008}. \bibinfo{publisher}{Springer},
  \bibinfo{pages}{46--57}.
\newblock
\showISBNx{978-3-319-16627-8}
\showDOI{%
\url{https://doi.org/10.1007/978-3-319-16628-5_4}}


\bibitem[\protect\citeauthoryear{Johansson}{Johansson}{1973}]%
        {J73}
\bibfield{author}{\bibinfo{person}{Gunnar Johansson}.}
  \bibinfo{year}{1973}\natexlab{}.
\newblock \showarticletitle{Visual perception of biological motion and a model
  for its analysis}.
\newblock \bibinfo{journal}{{\em Perception {\&} Psychophysics\/}}
  \bibinfo{volume}{14}, \bibinfo{number}{2} (\bibinfo{year}{1973}),
  \bibinfo{pages}{201--211}.
\newblock
\showISSN{1532-5962}
\showDOI{%
\url{https://doi.org/10.3758/BF03212378}}


\bibitem[\protect\citeauthoryear{Kamruzzaman and Begg}{Kamruzzaman and
  Begg}{2006}]%
        {KB06}
\bibfield{author}{\bibinfo{person}{J. Kamruzzaman} {and} \bibinfo{person}{R.~K.
  Begg}.} \bibinfo{year}{2006}\natexlab{}.
\newblock \showarticletitle{Support Vector Machines and Other Pattern
  Recognition Approaches to the Diagnosis of Cerebral Palsy Gait}.
\newblock \bibinfo{journal}{{\em IEEE Transactions on Biomedical
  Engineering\/}} \bibinfo{volume}{53}, \bibinfo{number}{12}
  (\bibinfo{date}{Dec} \bibinfo{year}{2006}), \bibinfo{pages}{2479--2490}.
\newblock
\showISSN{0018-9294}
\showDOI{%
\url{https://doi.org/10.1109/TBME.2006.883697}}


\bibitem[\protect\citeauthoryear{Kastaniotis, Theodorakopoulos, Theoharatos,
  Economou, and Fotopoulos}{Kastaniotis et~al\mbox{.}}{2015}]%
        {KTTEF15}
\bibfield{author}{\bibinfo{person}{Dimitris Kastaniotis},
  \bibinfo{person}{Ilias Theodorakopoulos}, \bibinfo{person}{Christos
  Theoharatos}, \bibinfo{person}{George Economou}, {and}
  \bibinfo{person}{Spiros Fotopoulos}.} \bibinfo{year}{2015}\natexlab{}.
\newblock \showarticletitle{{A Framework for Gait-Based Recognition Using
  Kinect}}.
\newblock \bibinfo{journal}{{\em Pattern Recognition Letters\/}}
  \bibinfo{volume}{68, Part 2} (\bibinfo{year}{2015}),
  \bibinfo{pages}{327--335}.
\newblock
\showISSN{0167-8655}
\showDOI{%
\url{https://doi.org/10.1016/j.patrec.2015.06.020}}


\bibitem[\protect\citeauthoryear{Kocsor, Kov{\'a}cs, and Szepesv{\'a}ri}{Kocsor
  et~al\mbox{.}}{2004}]%
        {KKS04}
\bibfield{author}{\bibinfo{person}{Andr{\'a}s Kocsor},
  \bibinfo{person}{Korn{\'e}l Kov{\'a}cs}, {and} \bibinfo{person}{Csaba
  Szepesv{\'a}ri}.} \bibinfo{year}{2004}\natexlab{}.
\newblock \bibinfo{booktitle}{{\em Margin Maximizing Discriminant Analysis}}.
\newblock \bibinfo{publisher}{Springer Berlin Heidelberg},
  \bibinfo{address}{Berlin, Heidelberg}, \bibinfo{pages}{227--238}.
\newblock
\showISBNx{978-3-540-30115-8}
\showDOI{%
\url{https://doi.org/10.1007/978-3-540-30115-8_23}}


\bibitem[\protect\citeauthoryear{Krzeszowski, Switonski, Kwolek, Josinski, and
  Wojciechowski}{Krzeszowski et~al\mbox{.}}{2014}]%
        {KSKJW14}
\bibfield{author}{\bibinfo{person}{Tomasz Krzeszowski}, \bibinfo{person}{Adam
  Switonski}, \bibinfo{person}{Bogdan Kwolek}, \bibinfo{person}{Henryk
  Josinski}, {and} \bibinfo{person}{Konrad Wojciechowski}.}
  \bibinfo{year}{2014}\natexlab{}.
\newblock \showarticletitle{{DTW-Based Gait Recognition from Recovered 3-D
  Joint Angles and Inter-ankle Distance}}. In \bibinfo{booktitle}{{\em Proc. of
  Computer Vision and Graphics: International Conference, ICCVG 2014, Warsaw,
  Poland}} {\em (\bibinfo{series}{LNCS})},
  \bibfield{editor}{\bibinfo{person}{Leszek~J. Chmielewski},
  \bibinfo{person}{Ryszard Kozera}, \bibinfo{person}{Bok-Suk Shin}, {and}
  \bibinfo{person}{Konrad Wojciechowski}} (Eds.), Vol.~\bibinfo{volume}{8671}.
  \bibinfo{publisher}{Springer}, \bibinfo{pages}{356--363}.
\newblock
\showISBNx{978-3-319-11331-9}
\showDOI{%
\url{https://doi.org/10.1007/978-3-319-11331-9_43}}


\bibitem[\protect\citeauthoryear{Kumar and Babu}{Kumar and Babu}{2012}]%
        {NV12}
\bibfield{author}{\bibinfo{person}{M.~S.~Naresh Kumar} {and}
  \bibinfo{person}{R.~Venkatesh Babu}.} \bibinfo{year}{2012}\natexlab{}.
\newblock \showarticletitle{Human Gait Recognition Using Depth Camera: A
  Covariance Based Approach}. In \bibinfo{booktitle}{{\em Proc. of the Eighth
  Indian Conference on Computer Vision, Graphics and Image Processing}} {\em
  (\bibinfo{series}{ICVGIP '12})}. \bibinfo{publisher}{ACM},
  \bibinfo{address}{New York, NY, USA}, Article \bibinfo{articleno}{20},
  \bibinfo{numpages}{6}~pages.
\newblock
\showISBNx{978-1-4503-1660-6}
\showDOI{%
\url{https://doi.org/10.1145/2425333.2425353}}


\bibitem[\protect\citeauthoryear{Kwolek, Krzeszowski, Michalczuk, and
  Josinski}{Kwolek et~al\mbox{.}}{2014}]%
        {KKMJ14}
\bibfield{author}{\bibinfo{person}{Bogdan Kwolek}, \bibinfo{person}{Tomasz
  Krzeszowski}, \bibinfo{person}{Agnieszka Michalczuk}, {and}
  \bibinfo{person}{Henryk Josinski}.} \bibinfo{year}{2014}\natexlab{}.
\newblock \showarticletitle{{3D Gait Recognition Using Spatio-Temporal Motion
  Descriptors}}. In \bibinfo{booktitle}{{\em Proc. of Intelligent Information
  and Database Systems: 6th Asian Conference, ACIIDS 2014, Bangkok, Thailand,
  Part II}} {\em (\bibinfo{series}{LNCS})}, Vol.~\bibinfo{volume}{8398}.
  \bibinfo{publisher}{Springer}, \bibinfo{pages}{595--604}.
\newblock
\showISBNx{978-3-319-05457-5}
\showDOI{%
\url{https://doi.org/10.1007/978-3-319-05458-2_61}}


\bibitem[\protect\citeauthoryear{Li, Jiang, and Zhang}{Li
  et~al\mbox{.}}{2006}]%
        {LJZ06}
\bibfield{author}{\bibinfo{person}{Haifeng Li}, \bibinfo{person}{Tao Jiang},
  {and} \bibinfo{person}{Keshu Zhang}.} \bibinfo{year}{2006}\natexlab{}.
\newblock \showarticletitle{{Efficient and Robust Feature Extraction by Maximum
  Margin Criterion}}.
\newblock \bibinfo{journal}{{\em IEEE Transactions on Neural Networks\/}}
  \bibinfo{volume}{17}, \bibinfo{number}{1} (\bibinfo{date}{Jan}
  \bibinfo{year}{2006}), \bibinfo{pages}{157--165}.
\newblock
\showISSN{1045-9227}
\showDOI{%
\url{https://doi.org/10.1109/TNN.2005.860852}}


\bibitem[\protect\citeauthoryear{Loog, Duin, and Haeb-Umbach}{Loog
  et~al\mbox{.}}{2001}]%
        {LDH01}
\bibfield{author}{\bibinfo{person}{Marco Loog}, \bibinfo{person}{Robert~P.W.
  Duin}, {and} \bibinfo{person}{Reinhold Haeb-Umbach}.}
  \bibinfo{year}{2001}\natexlab{}.
\newblock \showarticletitle{{Multiclass Linear Dimension Reduction by Weighted
  Pairwise Fisher Criteria}}.
\newblock \bibinfo{journal}{{\em IEEE Transactions on Pattern Analysis and
  Machine Intelligence\/}} \bibinfo{volume}{23}, \bibinfo{number}{7}
  (\bibinfo{date}{July} \bibinfo{year}{2001}), \bibinfo{pages}{762--766}.
\newblock
\showISSN{0162-8828}
\showDOI{%
\url{https://doi.org/10.1109/34.935849}}


\bibitem[\protect\citeauthoryear{Makihara, Mannami, and Yagi}{Makihara
  et~al\mbox{.}}{2011}]%
        {MMY10}
\bibfield{author}{\bibinfo{person}{Yasushi Makihara},
  \bibinfo{person}{Hidetoshi Mannami}, {and} \bibinfo{person}{Yasushi Yagi}.}
  \bibinfo{year}{2011}\natexlab{}.
\newblock \showarticletitle{Gait Analysis of Gender and Age Using a Large-scale
  Multi-view Gait Database}. In \bibinfo{booktitle}{{\em Proceedings of the
  10th Asian Conference on Computer Vision - Volume Part II}} {\em
  (\bibinfo{series}{ACCV'10})}. \bibinfo{publisher}{Springer-Verlag},
  \bibinfo{address}{Berlin, Heidelberg}, \bibinfo{pages}{440--451}.
\newblock
\showISBNx{978-3-642-19308-8}
\showURL{%
\url{http://dl.acm.org/citation.cfm?id=1965992.1966027}}


\bibitem[\protect\citeauthoryear{Murray}{Murray}{1967}]%
        {M67}
\bibfield{author}{\bibinfo{person}{M.~P. Murray}.}
  \bibinfo{year}{1967}\natexlab{}.
\newblock \showarticletitle{{Gait as a total pattern of movement: Including a
  bibliography on gait}}.
\newblock \bibinfo{journal}{{\em American Journal of Physical Medicine \&
  Rehabilitation\/}} \bibinfo{volume}{46}, \bibinfo{number}{1}
  (\bibinfo{year}{1967}), \bibinfo{pages}{290}.
\newblock


\bibitem[\protect\citeauthoryear{Preis, Kessel, Werner, and
  Linnhoff-Popien}{Preis et~al\mbox{.}}{2012}]%
        {PKWL12}
\bibfield{author}{\bibinfo{person}{Johannes Preis}, \bibinfo{person}{Moritz
  Kessel}, \bibinfo{person}{Martin Werner}, {and} \bibinfo{person}{Claudia
  Linnhoff-Popien}.} \bibinfo{year}{2012}\natexlab{}.
\newblock \showarticletitle{{Gait Recognition with Kinect}}. In
  \bibinfo{booktitle}{{\em 1st International Workshop on Kinect in Pervasive
  Computing, New Castle, UK, June 18--22}}. \bibinfo{pages}{1--4}.
\newblock
\showURL{%
\url{https://www.researchgate.net/publication/239862819_Gait_Recognition_with_Kinect}}


\bibitem[\protect\citeauthoryear{Sedmidubsky, Valcik, Balazia, and
  Zezula}{Sedmidubsky et~al\mbox{.}}{2012}]%
        {SVBZ12}
\bibfield{author}{\bibinfo{person}{Jan Sedmidubsky}, \bibinfo{person}{Jakub
  Valcik}, \bibinfo{person}{Michal Balazia}, {and} \bibinfo{person}{Pavel
  Zezula}.} \bibinfo{year}{2012}\natexlab{}.
\newblock \showarticletitle{{Gait Recognition Based on Normalized Walk
  Cycles}}.
\newblock In \bibinfo{booktitle}{{\em Advances in Visual Computing}}.
  \bibinfo{series}{LNCS}, Vol.~\bibinfo{volume}{7432}.
  \bibinfo{publisher}{Springer}, \bibinfo{pages}{11--20}.
\newblock
\showISBNx{978-3-642-33190-9}
\showDOI{%
\url{https://doi.org/10.1007/978-3-642-33191-6_2}}


\bibitem[\protect\citeauthoryear{Sinha, Chakravarty, and Bhowmick}{Sinha
  et~al\mbox{.}}{2013}]%
        {SCB13}
\bibfield{author}{\bibinfo{person}{Aniruddha Sinha}, \bibinfo{person}{Kingshuk
  Chakravarty}, {and} \bibinfo{person}{Brojeshwar Bhowmick}.}
  \bibinfo{year}{2013}\natexlab{}.
\newblock \showarticletitle{{Person Identification Using Skeleton Information
  from Kinect}}. In \bibinfo{booktitle}{{\em ACHI 2013: Proc. of the Sixth
  Intl. Conf. on Advances in CHI}}. \bibinfo{pages}{101--108}.
\newblock
\showURL{%
\url{https://www.thinkmind.org/index.php?view=article&articleid=achi_2013_4_50_20187}}


\bibitem[\protect\citeauthoryear{Su, Liao, and Chen}{Su et~al\mbox{.}}{2009}]%
        {SLC09}
\bibfield{author}{\bibinfo{person}{Han Su}, \bibinfo{person}{Zhi-Wu Liao},
  {and} \bibinfo{person}{Guo-Yue Chen}.} \bibinfo{year}{2009}\natexlab{}.
\newblock \showarticletitle{{A gait recognition method using L1-PCA and LDA}}.
  In \bibinfo{booktitle}{{\em {Proc. of the Eighth Intl. Conf. on Machine
  Learning and Cybernetics}}}, Vol.~\bibinfo{volume}{6}.
  \bibinfo{pages}{3198--3203}.
\newblock
\showISSN{2160-133X}
\showDOI{%
\url{https://doi.org/10.1109/ICMLC.2009.5212776}}


\bibitem[\protect\citeauthoryear{Tafazzoli, Bebis, Louis, and
  Hussain}{Tafazzoli et~al\mbox{.}}{2015}]%
        {TBLH15}
\bibfield{author}{\bibinfo{person}{Faezeh Tafazzoli}, \bibinfo{person}{George
  Bebis}, \bibinfo{person}{Sushil~J. Louis}, {and} \bibinfo{person}{Muhammad
  Hussain}.} \bibinfo{year}{2015}\natexlab{}.
\newblock \showarticletitle{{Genetic Feature Selection for Gait Recognition}}.
\newblock \bibinfo{journal}{{\em J. of Electron. Imaging\/}}
  \bibinfo{volume}{24}, \bibinfo{number}{1} (\bibinfo{date}{25 Feb.}
  \bibinfo{year}{2015}), \bibinfo{pages}{013036}.
\newblock
\showDOI{%
\url{https://doi.org/10.1117/1.JEI.24.1.013036}}


\bibitem[\protect\citeauthoryear{Vapnik}{Vapnik}{1995}]%
        {V95}
\bibfield{author}{\bibinfo{person}{Vladimir~N. Vapnik}.}
  \bibinfo{year}{1995}\natexlab{}.
\newblock \bibinfo{booktitle}{{\em The Nature of Statistical Learning Theory}}.
\newblock \bibinfo{publisher}{Springer-Verlag}, \bibinfo{address}{New York, NY,
  USA}.
\newblock
\showISBNx{0-387-94559-8}


\bibitem[\protect\citeauthoryear{Wahab and Bakar}{Wahab and Bakar}{2011}]%
        {WB11}
\bibfield{author}{\bibinfo{person}{Y. Wahab} {and} \bibinfo{person}{N.~A.
  Bakar}.} \bibinfo{year}{2011}\natexlab{}.
\newblock \showarticletitle{Gait analysis measurement for sport application
  based on ultrasonic system}. In \bibinfo{booktitle}{{\em 2011 IEEE 15th
  International Symposium on Consumer Electronics (ISCE)}}.
  \bibinfo{pages}{20--24}.
\newblock
\showISSN{0747-668X}
\showDOI{%
\url{https://doi.org/10.1109/ISCE.2011.5973775}}


\bibitem[\protect\citeauthoryear{Zeng and Wang}{Zeng and Wang}{2014}]%
        {ZW14}
\bibfield{author}{\bibinfo{person}{Wei Zeng} {and} \bibinfo{person}{Cong
  Wang}.} \bibinfo{year}{2014}\natexlab{}.
\newblock \showarticletitle{{View-Invariant Gait Recognition via Deterministic
  Learning}}. In \bibinfo{booktitle}{{\em {International Joint Conference on
  Neural Networks (IJCNN)}}}. \bibinfo{pages}{3465--3472}.
\newblock
\showISSN{2161-4393}
\showDOI{%
\url{https://doi.org/10.1109/IJCNN.2014.6889507}}


\end{thebibliography}

\end{document}